%% file: arxiv.tex
\documentclass[11pt, a4paper, logo, two column]{craftjarvis}

\pdftrailerid{redacted}

\makeatletter
\renewcommand\bibentry[1]{\nocite{#1}{\frenchspacing\@nameuse{BR@r@#1\@extra@b@citeb}}}
\makeatother

\input{config}

\usepackage[capitalize]{cleveref}

\usepackage[authoryear, sort&compress, round]{natbib} 

\newcommand{\method}{{\bf ActVLP}\xspace}
\newcommand{\model}{{\bf JARVIS-VLA}\xspace}

\title{
JARVIS-VLA: Post-Training Large-Scale Vision Language Models to Play Visual Games with Keyboards and Mouse
}

\reportnumber{} 

\renewcommand{\today}{March 2025} 

\author[1$\dag$]{Muyao~Li}
\author[1$\dag$]{Zihao~Wang}
\author[1]{Kaichen~He}
\author[2]{Xiaojian~Ma}
\author[1\textrm{\Letter}]{Yitao~Liang}

\affil[1]{Peking~University}
\affil[2]{BIGAI}
\affil[ \hspace{-0.73ex}]{All authors are affiliated with Team CraftJarvis} 

\begin{abstract} 
\input{abstract}
\end{abstract}

\begin{document}

\correspondingauthor{Yitao~Liang~<yitaol@pku.edu.cn>\\
$\dag$ indicates co-first author.\\
Muyao Li <2200017405@stu.pku.edu.cn>, Zihao Wang <zhwang@stu.pku.edu.cn>, Xiaojian Ma <xiaojian.ma@ucla.edu> 
}

\maketitle

\input{content_for_arxiv}

\section*{Limitations}

Looking ahead, there are several avenues for improvement in future work. First, it is crucial to enhance the inference throughput of \model, which is currently constrained by the large parameter size of the VLA based on VLM~\citep{EdgeVLA}. We believe that future integration with MoE~\citep{MoE,Switch_Transformers} could further improve the model’s inference efficiency, with the goal of achieving gameplay performance levels exceeding 40Hz. Additionally, there remains potential for further performance gains. While \model outperforms previous Minecraft policies, it still falls short of the performance demonstrated by top human players, who achieve success rates above 90\%.

\section*{Acknowledgement}

This work is funded in part by the National Science and Technology Major Project 2022ZD0114902.
We thank a grant from CCF-Baidu Open Fund.


\bibliographystyle{abbrvnat}
\bibliography{reference}

\newpage
\onecolumn
\appendix

\input{appendix}

\end{document}

%% file: config.tex
\usepackage{xcolor}
\definecolor{Gray}{gray}{0.9}
\definecolor{mygreen}{rgb}{0.0, 0.5, 0.0}
\definecolor{myred}{rgb}{0.8, 0.25, 0.33}
\definecolor{myblue}{rgb}{0.19, 0.55, 0.91}
\definecolor{uclablue}{rgb}{0.15, 0.45, 0.68}
\definecolor{boxgreen}{rgb}{0.02, 0.66, 0.02}
\definecolor{boxred}{rgb}{0.66, 0.1, 0.1}
\definecolor{boxblue}{rgb}{0.01, 0.01, 0.73}
\definecolor{mygray}{gray}{0.4}
\PassOptionsToPackage{numbers, compress}{natbib}
\usepackage{times}
\usepackage{cuted}
\usepackage{arydshln}
\usepackage{hyperref}
\usepackage{marvosym}
\usepackage{graphicx}
\usepackage{amssymb}
\usepackage{url}

\usepackage{microtype}
\usepackage{subfigure}
\usepackage{booktabs}

\usepackage{amsmath}
\usepackage{mathtools}
\usepackage{amsthm}

\usepackage{listings}
\usepackage{etoc}
\usepackage{algorithm}
\usepackage{algorithmic}
\usepackage{lipsum}
\usepackage{pifont}
\usepackage{wrapfig}
\usepackage{colortbl}
\usepackage{acronym}
\usepackage{multicol}
\usepackage{multirow}
\usepackage{makecell}
\usepackage{enumitem}
\usepackage{tabularx}
\usepackage{colortbl}
\usepackage{array}
\usepackage[skip=3pt,font=small]{caption}
\usepackage{xspace}
\usepackage{mdframed}

\usepackage[export]{adjustbox}

\newcolumntype{Y}{>{\arraybackslash}X}

\usepackage[utf8]{inputenc} 
\usepackage[T1]{fontenc}    
\usepackage{hyperref}       
\usepackage{amsfonts}       
\usepackage{nicefrac}       

\usepackage{mathtools}
\usepackage{amssymb}
\usepackage{amsthm}
\usepackage{autobreak}

\renewcommand{\algorithmicrequire}{\textbf{Input:}}
\renewcommand{\algorithmicensure}{\textbf{Output:}}

\renewcommand{\paragraph}[1]{\noindent\textbf{#1.}}

\renewcommand{\algorithmicrequire}{\textbf{Input:}}
\renewcommand{\algorithmicensure}{\textbf{Output:}}

\renewcommand{\paragraph}[1]{\noindent\textbf{#1.}}

\makeatletter
\DeclareRobustCommand\onedot{\futurelet\@let@token\@onedot}
\def\@onedot{\ifx\@let@token.\else.\null\fi\xspace}

\makeatother

\acrodef{llms}[LLMs]{Large Language Models}
\acrodef{mlms}[MLMs]{Multimodal Language Models}

\usepackage{xcolor}

\usepackage[most]{tcolorbox}
\newtcolorbox[list inside=prompt,auto counter,number within=section]{prompt}[1][]{
    colbacktitle=black!60,
    coltitle=white,
    fontupper=\footnotesize,
    boxsep=5pt,
    left=0pt,
    right=0pt,
    top=0pt,
    bottom=0pt,
    boxrule=1pt,
    #1,
}

%% file: abstract.tex
Recently, action-based decision-making in open-world environments has gained significant attention. Visual Language Action (VLA) models, pretrained on large-scale web datasets, have shown promise in decision-making tasks. However, previous work has primarily focused on action post-training, often neglecting enhancements to the foundation model itself. 
In response, we introduce Act from Visual Language Post-Training (\method), a novel training paradigm. \method distinctively enhances the foundation model prior to action-specific tuning by first post-training it on a curated set of environment-specific visual and linguistic tasks using self-supervised learning. This initial stage significantly improves the model's capabilities in world knowledge, visual recognition, and spatial grounding. 
Subsequently, this strengthened VLM undergoes action post-training via imitation learning on trajectory datasets.
Following this paradigm, we develop \model, the first VLA model in Minecraft that can follow human instructions on over 1k different atomic tasks, including crafting, smelting, cooking, mining, and killing. 
Our experiments demonstrate that our \method paradigm leads to a significant 40\% improvement over the best agent baseline on a diverse set of atomic tasks. Furthermore, \model surpasses traditional imitation learning-based policies in Minecraft, achieving state-of-the-art performance. We have open-sourced the code, models, and datasets to foster further research.
The project page can be found at \url{https://craftjarvis.github.io/JarvisVLA}.

%% file: content_for_arxiv.tex
\section{Introduction}

\begin{figure*}[t!]
    \centering
    \includegraphics[width=0.95\linewidth]{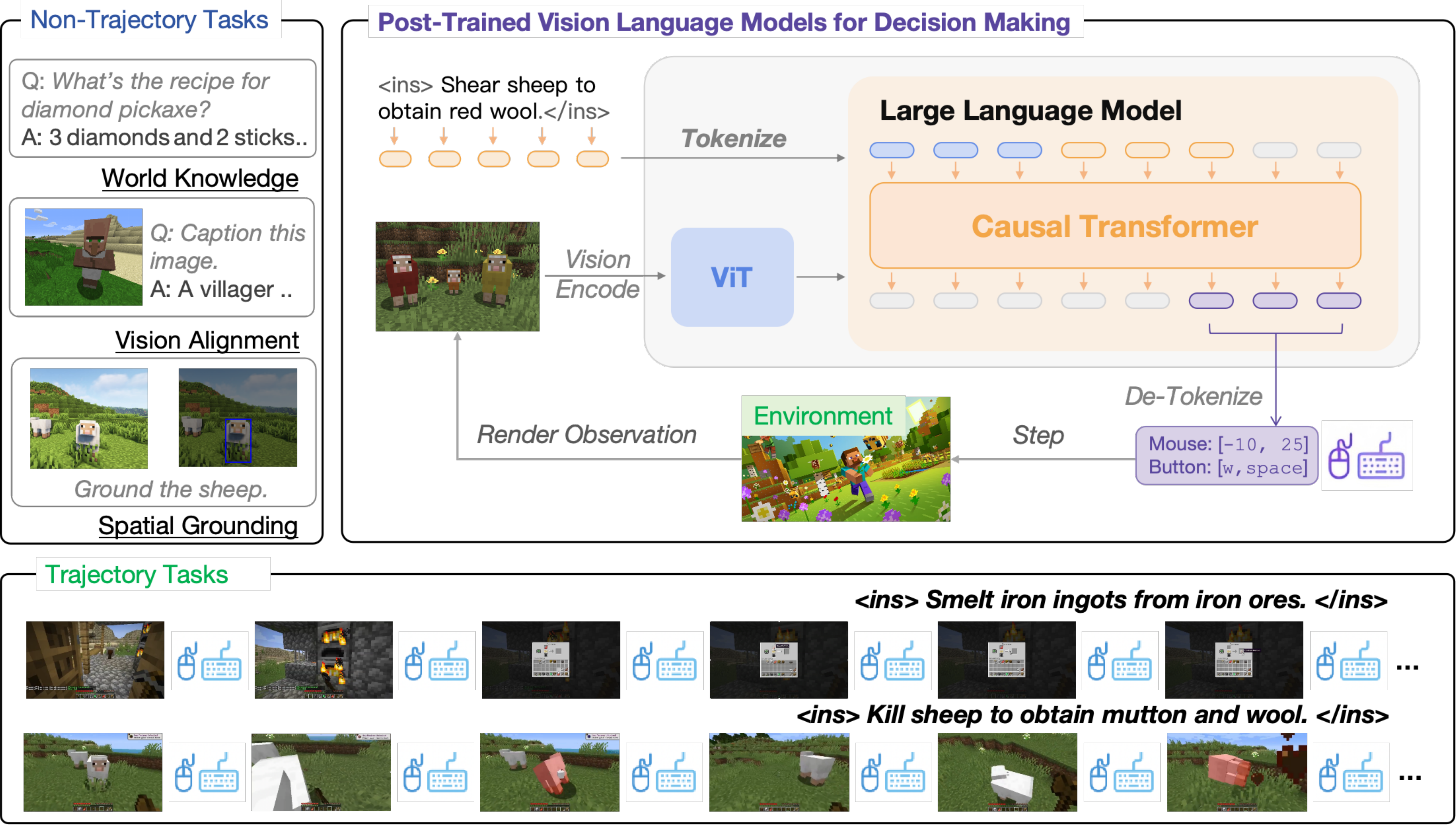}
    \caption{We present \model, a novel Vision-Language-Action (VLA) model trained with \method paradigm, post-trained on vision language tasks (non-decision-making tasks) before training on trajectory datasets to have better decision-making capabilities.
    }
    \label{fig:teaser}
\end{figure*}

Pretraining foundation models on large-scale, noisy internet datasets has become a mainstream approach in NLP and vision~\citep{vit,gpt-4,qwen2vl,gemini}. The success of models like GPT and LLAMA~\citep{llama,chatgpt} has shown that large, capable language models can infer and execute tasks described by language prompts. However, this paradigm has yet to achieve similar success in the decision-making domain~\citep{yang2023foundation,cheng2024exploring}. In particular, while OpenAI's Video Pre-Training (VPT) model~\citep{vpt} has attempted to apply a similar approach in Minecraft, it still relies heavily on imitation learning (IL) after collecting large-scale YouTube videos of human play. VPT's approach of pretraining with imitation learning, followed by downstream supervised fine-tuning and reinforcement learning, made significant strides—culminating in the successful \texttt{ObtainDiamond}, a key challenge in Minecraft\footnote{Diamond tools are considered a grand challenge, with experienced human players taking up to 20 minutes (24,000 actions) to craft them.}.

Despite this success, the reliance on next-action prediction in imitation learning limits the development of robust, multi-task decision-making abilities~\citep{rt-1,gr1,octo,openx}. Moreover, this pretraining paradigm struggles to generalize to unseen environments or tasks due to the intricacies of the interactions between observations and behavior, whereas language tokens are more standardized. To overcome these challenges, a new approach has emerged that leverages pretrained Vision Language Models (VLMs) for decision-making. These models, known as Vision Language Action models (VLAs), integrate language understanding with action generation and can be further enhanced through post-training on visual-language tasks~\citep{openVLA,3d-vla}. A more detailed discussion can be found in \autoref{fig:teaser}~(left) and \autoref{sec:training_pipeline}.

However, much like traditional imitation learning, current VLA approaches predominantly focus on action post-training. In these models, the learning objective is to generate correct actions based on large-scale cross-task imitation data. We propose that, in addition to action generation, understanding the environment and incorporating task-related knowledge could be equally important for achieving more flexible and generalizable decision-making. To this end, we introduce a novel training paradigm—Vision Language Post-Training (\method)—which integrates visual-language tasks into the post-training phase of VLA models. Following the above paradigms, we obtain the first VLA models in Minecraft that can follow human instructions on over 1k different atomic tasks, including crafting, smelting, cooking, mining, and killing. 

Our contributions are as follows: (1) We pioneer the use of VLA in the open-world environment of Minecraft by introducing \model, a powerful model achieving state-of-the-art performance in action-based decision-making. 
(2) We introduce the concept of Vision Language Post-Training and identify key visual-language guidance strategies that enhance decision-making. (3) We investigate the scaling laws of VLA models, demonstrating that expanding the scale of non-trajectory vision-language tasks during post-training leads to significant improvements in downstream task performance. (4) We open-source the code, models, and datasets to support further research in this area.

\begin{figure*}[t!]
    \centering
    \includegraphics[width=0.95\linewidth]{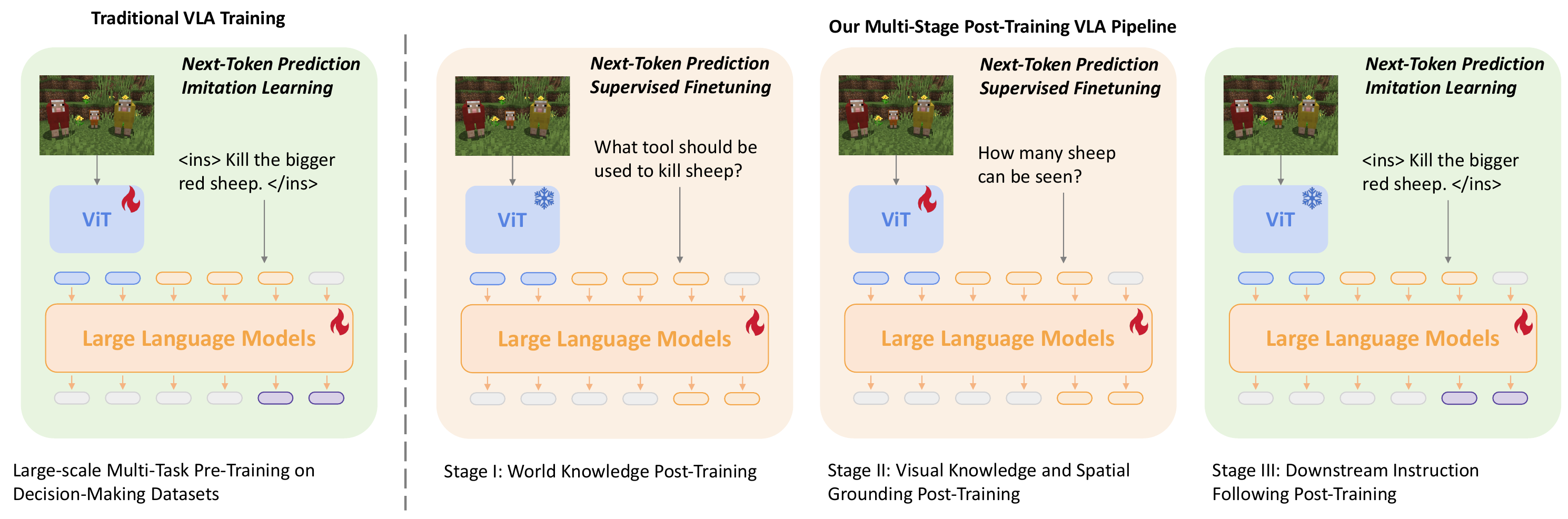}
    \caption{Previous VLA methods usually directly use imitation learning to finetune original vision-language models on large-scale multi-domain decision-making datasets to predict the actions~\citep{openVLA,rt-2}. Our \method training pipeline includes three stages: 1) post-training language models on text-only world knowledge with next-token prediction supervised fine-tuning, 2) post-training both vision encoder and language models on multimodal vision-language alignment and spatial grounding datasets with next-token prediction supervised fine-tuning, and 3) post-training only language models on multi-modal instruction following datasets with imitation learning. 
    }
    \label{fig:training_pipeline}
\end{figure*}

\section{Learning to Act from Vision Language Post-Training}

In this section, we present a detailed introduction to \method, a new paradigm for training VLA models. One of the most significant improvements is that we investigate a post-training stage prior to imitation learning. Specifically, we instantiate this paradigm in our proposed model, \model. We begin by discussing the architecture for \model in \autoref{sec:architecture}, followed by an explanation of the training pipeline in \autoref{sec:training_pipeline} and the datasets used in \autoref{sec:datasets}. 

\subsection{Model Structure}\label{sec:architecture}

As illustrated in \autoref{fig:teaser}, \model employs an architecture similar to Llava~\citep{llavanext} but with slight modifications. The structural framework,  consists of several key components:: 1) Visual Encoder: A Vision Transformer~\citep{vit} that processes raw image pixels and converts them into a sequence of fixed-size image patches. 2) Image Projection Module: A lightweight two-layer MLP that projects image patch embeddings into the same representational space as word embeddings. 3) Language Model Transformers~\citep{llama,qwen}: A powerful autoregressive language model that serves as the core of the system, facilitating multimodal reasoning and decision-making.

Unlike OpenVLA~\citep{openVLA}, our framework is designed for partially observable environments. To accommodate this, we adopt a non-Markovian architecture by incorporating a history of observation images within the prompt. This approach ensures that the model retains temporal context, which is crucial for tasks requiring multi-step reasoning and long-horizon decision-making. In our experiments, we employ Llava-Next~\citep{llavanext} and Qwen2-VL~\citep{qwen2vl} as base vision language models, as both models provide robust support for multi-image understanding, enabling enhanced perception and reasoning.

Another key distinguishing feature of \model compared to prior VLA models is the integration of an action decoder. This module is responsible for decoding both discrete and continuous actions. For discrete actions, we consolidate related action dimensions into unified categories to reduce redundancy and improve efficiency. For continuous actions, we discretize the action space into bins, which are then mapped to discrete tokens. These tokens are subsequently appended to the vocabulary of the original foundation model, allowing the model to generate both textual and action-based outputs in a unified manner.

Instead of retraining the base VLM's tokenizer, we adopt a strategy inspired by RT-2~\citep{rt-2}, repurposing the least frequently used tokens from the language tokenizer's vocabulary to represent action semantics. Specifically, we replace the 51 least-used tokens, allocating 22 tokens for mouse control (e.g., cursor movements) and 29 tokens for special keyboard inputs (e.g., function keys and command shortcuts), which can be found in \autoref{app:action_space}. 
We introduce no other modifications to the original VLM architecture to maintain model generalizability and ensure broad compatibility across different foundation models. This design choice allows \model to be easily integrated with various pre-trained multimodal models while preserving their inherent capabilities.

\begin{figure*}[t!]
    \centering
    \includegraphics[width=0.95\linewidth]{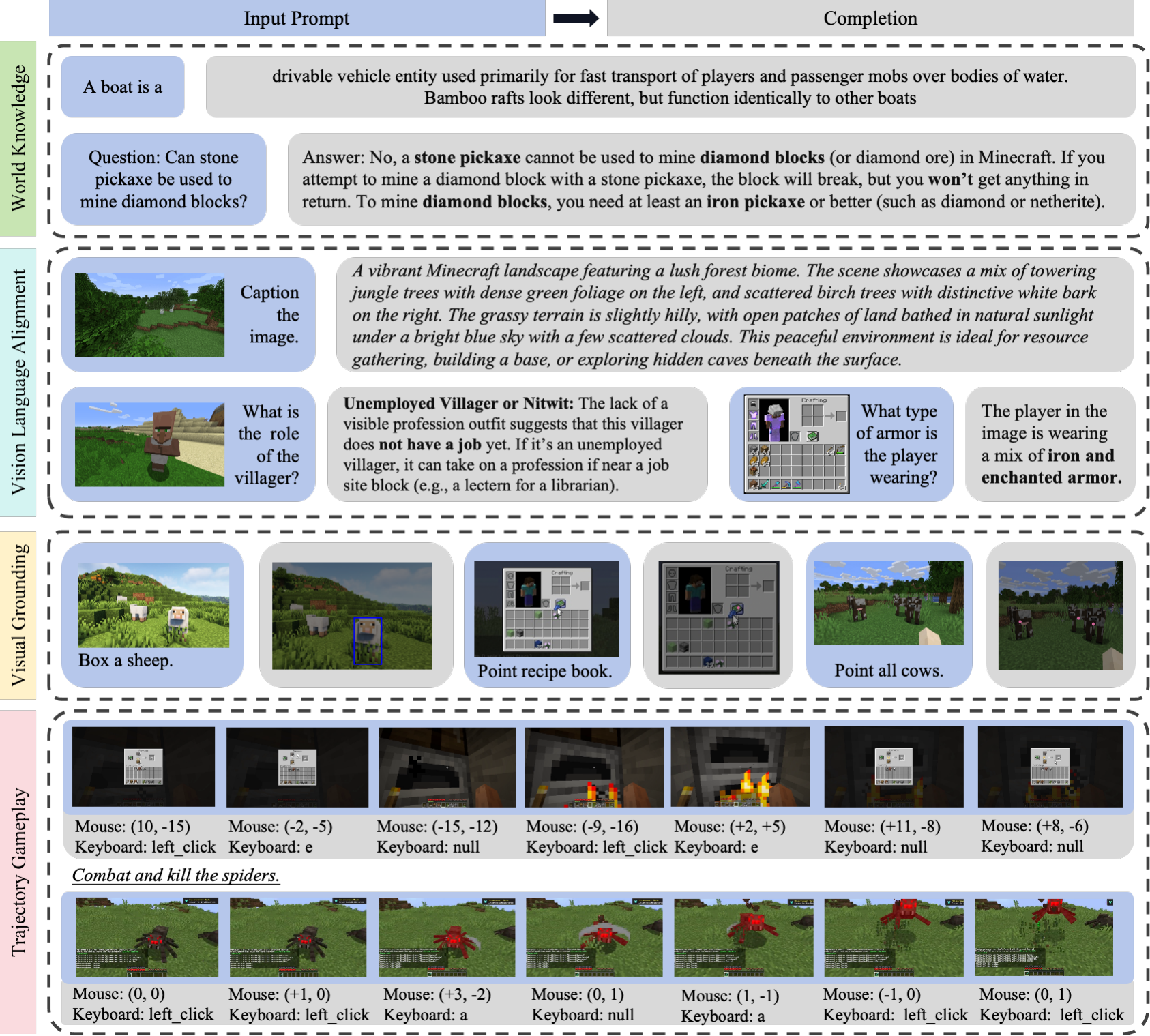}
    \caption{\textbf{Illustration of various post-training datasets.} 
    Models can post-train on various vision-language datasets using a unified tokenizer and support diverse vision-language applications, such as question answering, image captioning, image/video question answering, visual grounding (including points and bounding box), and decision-making.
    More examples can be found in \autoref{app:training_datasets}.
    }
    \label{fig:datasets}
\end{figure*}

\subsection{Training Pipeline}\label{sec:training_pipeline}

Traditional VLA methods typically employ pretrained VLMs and train them via imitation learning on large-scale trajectory data, which includes textual instructions, visual observations, and action token sequences, as illustrated in \autoref{fig:training_pipeline}(left). These methods assume that VLMs, pretrained on diverse internet-scale data, possess strong generalization and fitting capabilities. Consequently, they are fine-tuned directly on downstream decision-making tasks, leveraging multi-scenario data to enhance action understanding and generalization.

However, learning world knowledge from action-labeled trajectory data is inherently challenging~\citep{vpt}. Moreover, the lack of large-scale action-labeled datasets makes it challenging to pretrain expansive models using only trajectory data~\citep{openx}.

To address these challenges, \method enhances the VLM through a structured post-training process, utilizing data that follows the same format as pretraining but is more relevant to decision-making tasks. As shown in \autoref{fig:training_pipeline}(right), our training pipeline consists of three stages.

\paragraph{Stage I: Post-Training Language Models} We first refine the language transformer of the VLM using large-scale textual datasets related to world knowledge in downstream environments, e.g., Minecraft. During this stage, vision-related components, including the ViT and vision adapter modules, are frozen. This step enhances the model’s understanding of decision-making contexts before incorporating multimodal alignment.

\paragraph{Stage II: Post-Training Vision Encoder and Language Models} Following language post-training, we fully unfreeze the VLM and fine-tune it using captioning, visual question-answering (VQA), and spatial grounding datasets, which are multimodal and have images in datasets. This stage ensures improved vision-language alignment, enhancing the model’s capacity to integrate world knowledge with visual perception. Both Stage 1 and Stage 2 employ next-token prediction through supervised fine-tuning, with the optimization objective being:
\begin{equation}
\mathcal{L}_\text{SFT} = - \sum_{i=1} \log \mathcal{P}_\theta (x_i \mid x_v, x_{\text{ins}}, x_{1:i-1}) \tag{1}
\end{equation}
where $x_v$ denotes visual tokens, $x_{\text{ins}}$ represents the instruction, and $x$ corresponds to the answer. This loss function maintains consistency with the standard causal mask training approach.

\paragraph{Stage III: Action Post-Training for Interaction} The final stage of our pipeline is action post-training, where the VLM transitions from a passive observer to an active agent. During this action post-training, the VLM is fine-tuned on trajectory data ($\mathcal{D}$). The model learns to map textual instructions ($x_{\text{ins}}$) and current visual observations ($o_t \in \mathbb{R}^{H\times W\times 3}$) to action chunks $a_{t:t+\tau}$. The imitation learning objective is to maximize the likelihood of these expert actions:
\begin{equation}
\label{equation:3}
\mathcal{L}_\text{IL} = - \sum_{t=1} \log \pi_\theta (a_{t:t+\tau} \mid o_t, x_{\text{ins}}) \tag{2}
\end{equation}
where $\pi_\theta$ is the learned policy parameterized by $\theta$. We employ action chunking as this technique promotes temporally coherent and improves training efficiency~\citep{openvla-oft}. In this stage, the vision-related modules remain frozen. while the language tokenizer is modified to incorporate action tokens,
and the language transformer undergoes full parameter fine-tuning.

This structured pipeline ensures that the VLM is progressively refined before being adapted to trajectory-based imitation learning, resulting in improved world knowledge acquisition, vision-language alignment and grounding, and action generalization in decision-making tasks.

\subsection{Datasets}
\label{sec:datasets}

To support the \method training pipeline, we constructed a large-scale multimodal dataset. This dataset includes both non-trajectory task datasets for post-training and trajectory datasets for downstream imitation learning. The non-trajectory datasets are divided into three categories: knowledge-based question answering, vision language alignment, and spatial grounding. These categories are designed to enhance the model’s decision-making capabilities before trajectory fine-tuning. For trajectory datasets, we collected over 7.4 million frames of Minecraft gameplay data, including expert actions from diverse sources such as human-playing~\citep{vpt}, youtube videos, and existing agents~\citep{jarvis1}.

The dataset for \textit{world knowledge comprehension} comprises approximately 277K entries that significantly bolster textual understanding, employed during training stage I. The \textit{visual-language alignment} dataset incorporates 35K keyframes enhanced with advanced Vision-Language Models to produce both captions and question-answer pairs, facilitating multimodal supervised fine-tuning in the subsequent training stage. The \textit{spatial grounding} dataset focuses on detailed object localization, generating over 404K data points that are instrumental in refining spatial understanding for \method models. Both the visual-language alignment datasets and the spatial grounding datasets primarily utilize Minecraft observations, which strengthen the VLM's understanding of the world and are used to support training stage II. 

\paragraph{Imitation Learning Trajectory Dataset}
VLA training is constructed on a dataset of human gameplay trajectories, particularly from the OpenAI contractor dataset in Minecraft~\citep{vpt}, which includes diverse tasks. We also incorporated an additional 3M rollout frames from VPT~\citep{vpt} and JARVIS-1~\citep{jarvis1} agents. For structured GUI-based tasks like crafting and smelting, we synthesized 6.4M expert data entries to improve imitation learning. This trajectory data was partitioned into two distinct subsets. First, over 100 random trajectories for each MCU benchmark task were allocated for targeted fine-tuning. The remaining data, approximately 10 billion tokens, was then utilized for action post-training. Representative examples of our datasets are shown in \autoref{fig:datasets}, with further details in \autoref{app:training_datasets}.

\section{Experiments}

Our experiments (starting from subsection 4.2) aim to address the following questions:

\noindent \textbf{Q1}: How do \model compare to \textit{sota} open-world agents and imitation learning methods? 

\noindent \textbf{Q2}: Is vision language post-training the true cause of the performance improvement?

\noindent \textbf{Q3}: Whether VLAs exhibit scaling laws and how \method influences them?

\noindent \textbf{Q4}: Is \method sensitive to different VLM backbones? 
Due to space constraints, we quickly respond with an affirmative no, detailed experiment discussion deferred to \autoref{sec:experiments_vlm_ablation}.

\subsection{Experimental Setup}\label{sec:experimental_setups}

\paragraph{Evaluation Environment} 
We use Minecraft 1.16.5 as our experimental platform~\citep{minerl}. As an open-world game with a substantial knowledge base on platforms such as Reddit and wiki~\citep{minedojo}, Minecraft poses significant challenges to agents while simultaneously offering rich resources for research. To ensure fair comparisons, we align the action and visual observation spaces with those of human players~\citep{vpt}.
Additionally, we hide information unavailable to human players as well, such as agent location and inventory stats.

\input{tables/atom_tasks}

\paragraph{Benchmark and Evaluation Metrics}
We conduct evaluations using two broad benchmarks: (i) the agent’s capacity to interact with the Minecraft environment to complete tasks; and (ii) vision-language tasks (e.g., question answering, spatial grounding) designed to assess the VLM's understanding of Minecraft-specific knowledge.
For the instruction-following tasks, we adopt the MCU Benchmark~\citep{mcu}, focusing on four categories—\texttt{Mine Blocks}, \texttt{Kill Entities}, \texttt{Craft Items}, and \texttt{Smelt Items}—that represent a wide range of typical game-play behaviors in Minecraft. Notably, \texttt{Craft} and \texttt{Smelt} require 2D GUI manipulation through the mouse (covering thousands of item categories), whereas \texttt{Mine} and \texttt{Kill} involve recognizing, navigating, and interacting with targets in a 3D environment.
Each category contains at least 5 distinct tasks. For instance, the \texttt{Mine Blocks} category includes mining iron ore \includegraphics[scale=0.035,valign=c]{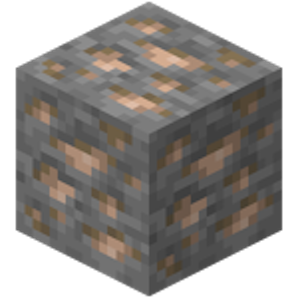} with a stone pickaxe, oak logs \includegraphics[scale=0.035,valign=c]{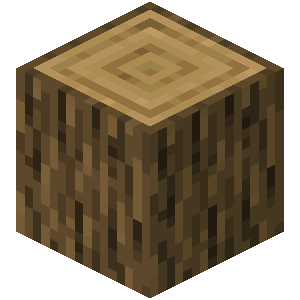} with bare hands, grass \includegraphics[scale=0.035,valign=c]{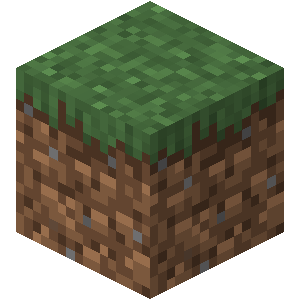}, dirt \includegraphics[scale=0.16,valign=c]{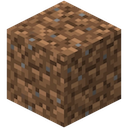}, and obsidian \includegraphics[scale=0.030,valign=c]{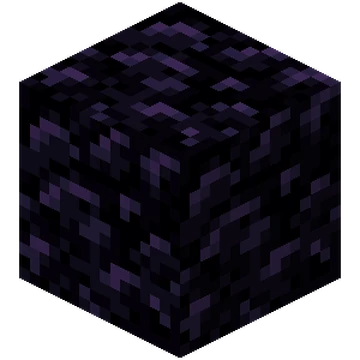} with a diamond pickaxe. Our evaluation set includes both simpler tasks (e.g., mining oak logs) and more complex ones (e.g., mining obsidian for over 10 seconds) that have proven challenging for prior state-of-the-art agents~\citep{groot,steve1}.
We perform each task at least 30 times and report the success rate per task, as well as the average success rate within each category. To ensure fairness, maximum execution steps for selected tasks match those reported by \citet{mcu}.
For vision-language assessments, the task formulations are illustrated in \autoref{fig:datasets}. We provide human-written ground-truth answers and employ an LLM-as-judge to evaluate the performance of various VLMs (GPT-4o, Llava, Qwen-VL, and our post-trained VLMs). Detailed information on these vision-language benchmarks and results can be found in \autoref{app:benchmarks}.

\paragraph{Training and VLA Configurations}. 
Our training pipeline follows the process described in \autoref{sec:training_pipeline}: we first obtain a visual-language post-training intermediate model, then further train it on trajectory tasks to produce the \model. We conduct experiments using two popular frameworks: Qwen2-VL~\citep{qwen2vl} and Llava~\citep{llavanext}. 
We develop a discretized action tokenizer specific to Minecraft, comprising 51 tokens that represent camera movements and button actions. 
We utilize the \texttt{trl} SFT Trainer~\citep{trl} for finetuning and deploy the VLA with \texttt{vLLM}~\citep{vllm}. Training is carried out on 32 A800-80G GPUs, while inference runs on a single NVIDIA RTX 3090. Further training details are provided in \autoref{app:training_configs}.

\paragraph{Baselines} 
We compare our model with: 1) VPT~\citep{vpt}, including both the behavior cloning (VPT-BC) and reinforcement learning (VPT-RL) variants; 2) STEVE-1\citep{steve1}, a text-conditioned policy that combines VPT and MineCLIP\citep{minedojo} for instruction following;
3) GROOT~\citep{groot}, which uses video prompts as task instructions; and 4) MineDreamer~\citep{minedreamer}, which leverages a vision-language model and a diffusion model to guide the STEVE-1 policy.
Each method follows the default configuration provided in the MCU benchmark for a fair comparison.

\begin{figure*}[t!]
    \centering
    \includegraphics[width=0.95\linewidth]{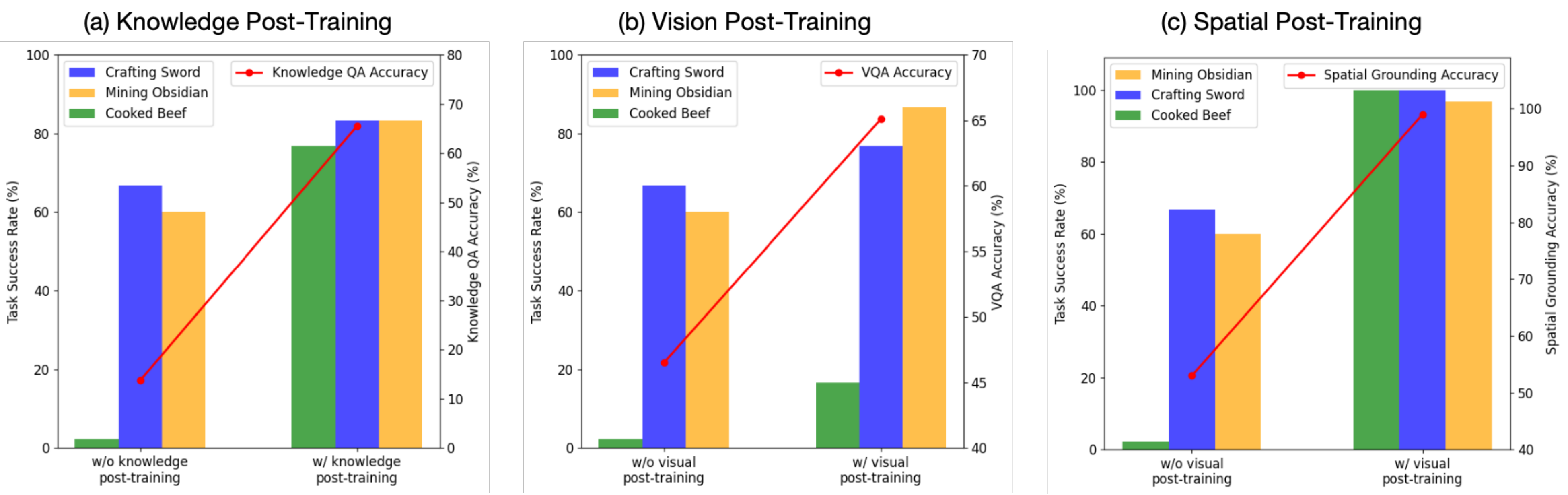}
    \caption{\textbf{Ablation results on different post-training datasets.} 
    We select knowledge datasets, visual question-answering datasets, and spatial grounding datasets to conduct ablation experiments. Our goal is to evaluate which capabilities and post-training datasets most significantly influence downstream decision-making tasks.
    }
    \label{fig:posttrain_dataset_ablation}
\end{figure*}

\subsection{VLA Performance Evaluation}
\label{sec:main_experiments}

We present the performance results of our proposed model across four categories from the MCU benchmark~\citep{mcu}, as shown in \autoref{tab:atom_tasks}. For each MCU task, we collect over 100 random trajectories, which are used to fine-tune base VLMs to create our final VLA models.

We evaluate three variants of the VLMs as base models:
1) Qwen2-VL~(raw): the original VLM checkpoint, fine-tuned directly on the task-specific dataset. 2) Qwen2-VL~(IL): first undergoes action post-training, then trained on the same task-specific fine-tuning dataset. 3) \model-Qwen2-VL: first post-trained using our proposed \method paradigm, then fine-tuned on the same dataset. Performance is measured by the average success rate across tasks within each category. Our results show that \model-Qwen2-VL, post-trained using our approach, consistently outperforms prior methods across almost all tasks.

Remarkably, even without task-specific post-training, raw Qwen2-VL model, fine-tuned on downstream tasks, outperforms several previous baselines, including STEVE-1~\citep{steve1} and GROOT~\citep{groot}, which were trained using large-scale imitation learning. This highlights the effectiveness of using a robust pre-trained VLM as the base model for the policy, leading to strong performance even without additional fine-tuning.

Notably, we observe a significant performance boost with \method post-training. For tasks such as \texttt{Craft Items} and \texttt{Smelt Items}, where previous methods struggled, \model-Qwen2-VL achieves success rates more than double those of the baseline models. This underscores the effectiveness of our off-trajectory vision-language task strategy. Furthermore, \model-Qwen2-VL outperforms Qwen2-VL~(IL) by over 15\%, despite using only 21\% of the training trajectory data. 
In crafting category tasks, the \model model surpasses traditional baselines by more than double, outperforming models like VPT-BC~\citep{vpt} and STEVE-1~\citep{steve1} on tasks such as "Craft crafting table" (\includegraphics[scale=0.08,valign=c]{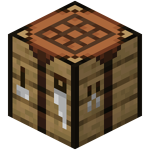}). This significant improvement is primarily due to the use of ViT in VLM and high-resolution processing, which are crucial for tasks like crafting and smelting that demand precise control in the GUI interface.
This suggests that integrating off-trajectory vision-language tasks into the training pipeline enhances decision-making capabilities, enabling more accurate action predictions in VLA models. 
Further analysis and additional experiments will be presented in the next section.

\subsection{Ablation on Training Paradigm}
\label{sec:experiment_paradigm}
To assess the impact of visual-language post-training, we designed an additional baseline, Qwen2-VL-7B (one-stage). It is co-trained on the original checkpoint of Qwen2-VL, using the combination of non-trajectory datasets and trajectory datasets in a single unified training stage. After training for one epoch, Qwen2-VL-7B (one-stage) is fine-tuned on MCU tasks. We compare it with \model-Qwen2-VL, which is post-trained in three separated steps.

\input{tables/paradigm_ablation}

The results presented in \autoref{tab:paradigm_ablation} indicate that separating visual-language post-training from action learning, rather than merging all stages into a single training phase, yields a significant performance boost. For instance, when comparing our \model-Qwen2-VL to the one-stage co-finetuning approach, we observed substantial improvements, including approximately 57\% on the "Smelt iron ingot \includegraphics[scale=0.04,valign=c]{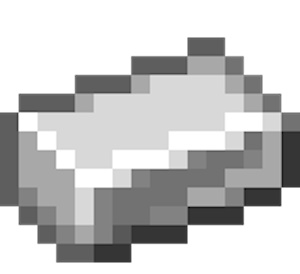}" task and 23\% on the "Crafting the ladder \includegraphics[scale=0.06,valign=c]{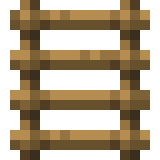}" task. We hypothesize that dedicated post-training on visual-language datasets enhances the foundational capabilities of the VLM, providing a more advantageous starting point for subsequent imitation learning.

\subsection{Ablation on Non-Trajectory Datasets}\label{sec:experiment_datasets_ablation}

In this section, we focus on the post-training of Qwen2-VL using various non-trajectory vision-language tasks to investigate the specific contributions to its enhanced performance.

To understand the impact of different task enhancements, we conduct an ablation study by dividing the non-trajectory datasets and training Qwen2-VL separately on three types of tasks: spatial grounding, vision language alignment, and knowledge-based question-answering, which are all related to Minecraft games. This results in three variants of the VLM, each augmented with one of these capabilities—spatial grounding, visual recognition, and world knowledge. All models are finetuned using the same gameplay dataset and imitation learning techniques. We also develop a benchmark, detailed in \autoref{app:benchmarks}, to evaluate these capabilities. For this evaluation, we select three long-sequence atomic tasks: "Craft the diamond sword" (\includegraphics[scale=0.08,valign=c]{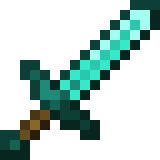}), "Mine the obsidian" (\includegraphics[scale=0.035,valign=c]{figures/avatar/obsidian.png}), and "Cook the beef" (\includegraphics[scale=0.038,valign=c]{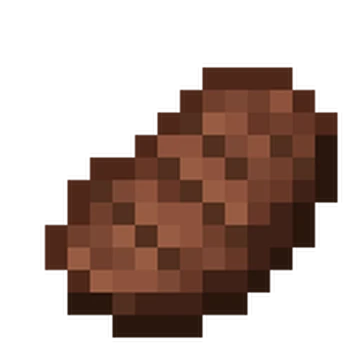}), as downstream instruction-following tasks.

The results of our ablation studies, presented in \autoref{fig:posttrain_dataset_ablation}, demonstrate that post-training with non-trajectory vision-language tasks significantly enhances the core capabilities of the VLM across the respective benchmarks. Notably, after fine-tuning, models enhanced with spatial grounding exhibit the most substantial improvement in downstream decision-making tasks. These findings underscore the effectiveness of non-trajectory post-training in boosting the performance of Vision-Language-Action models in decision-making tasks, even when the focus is on a single task. We find that non-trajectory vision-language tasks, which are essential for agent pipelines~\citep{deps,jarvis1}, are more effective for fine-tuning end-to-end VLA models. This demonstrates the connection between developing LLM-based agent pipelines with separate modules and fine-tuning end-to-end VLA models.

\begin{figure*}[t]
\centering
    \includegraphics[width=0.95\linewidth]{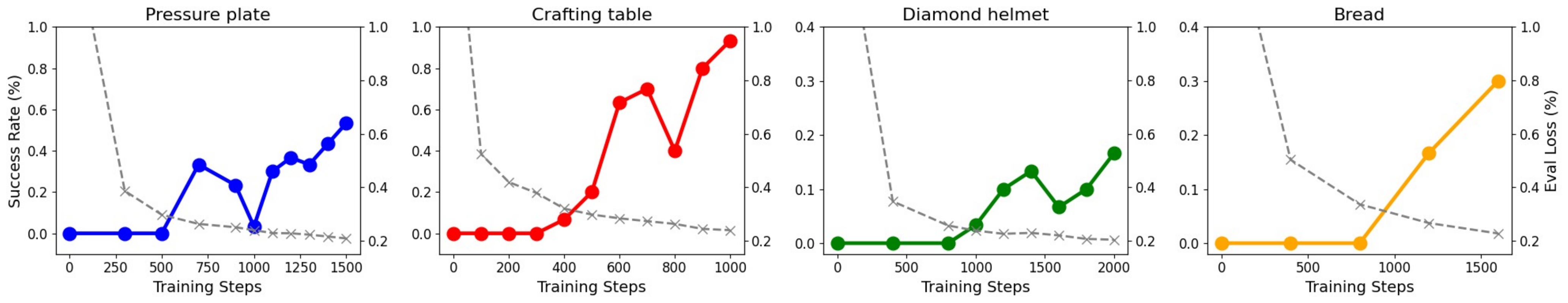}
    \caption{\textbf{The relation between downstream task success rate, training loss, and training steps.} The curve shows that scaling downstream finetuning trajectories can scale up the success rate when the loss is lower than 0.22.}
    \label{fig:scaling}
\end{figure*}

\begin{figure}
\centering
    \includegraphics[width=0.99\linewidth]{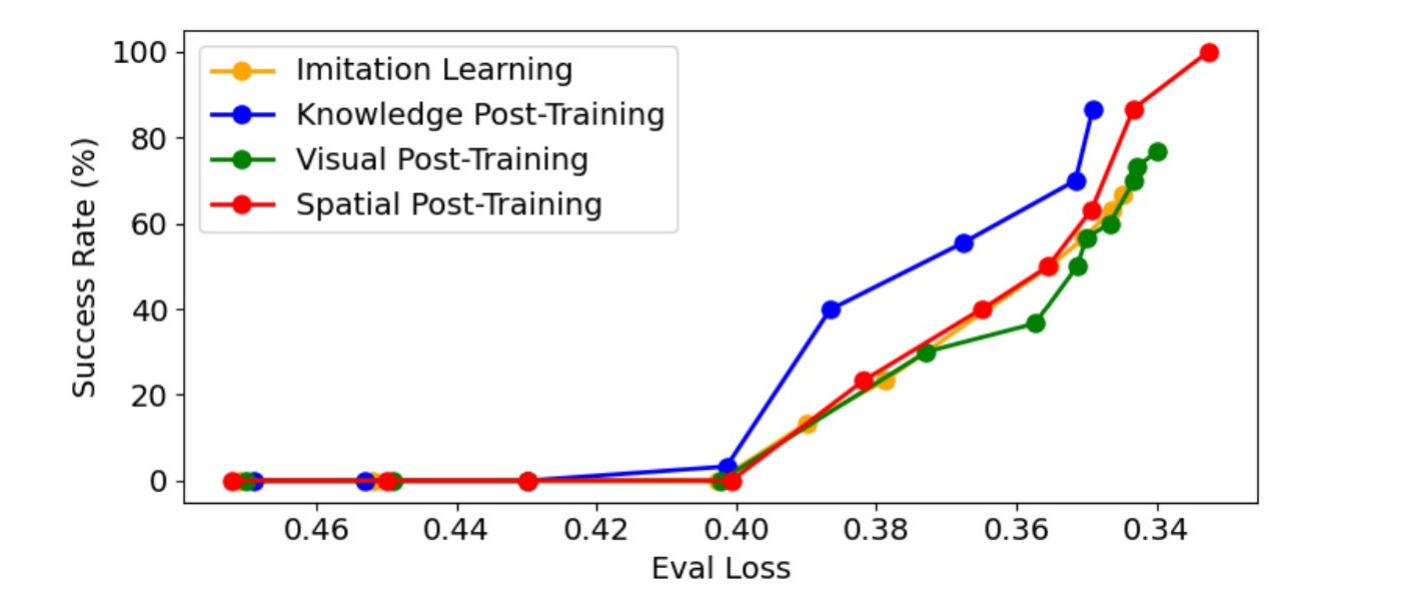}
    \caption{
    \textbf{The relationship between post-training loss and downstream task success rates.} Our findings indicate that increasing the size of post-training non-trajectory datasets can significantly enhance downstream task success rates, even with a fixed number of fine-tuning trajectories.
    }
    \label{fig:scaling_law_v2}
\end{figure}

\subsection{Scaling Experiments}\label{sec:more_analysis}

Recent work on large language models (LLMs) trained on vast amounts of text via next-token prediction has shown strong scaling laws~\citep{Emergent_abilities_openai,Emergent_abilities_zhipu,lin2024selecting,wangoptimizing}. We investigate whether VLAs, obtained through post-training on VLMs, exhibit similar scaling behavior. Specifically, we explore two questions: Q1) Can scaling up downstream imitation learning trajectories further improve the VLA’s task success rate? Q2) Does increasing the amount of non-trajectory vision-language tasks used during post-training enhance task completion success?

The results for Q1 are shown in \autoref{fig:scaling}. Using the same base model, we observe that increasing the number of downstream trajectories improves the VLA model’s task success rate. However, since the success rate is a discrete metric, we find that tasks only show a non-zero success rate when the evaluation loss is below 0.30. This indicates that the dataset size for downstream fine-tuning must be sufficiently large enough. Furthermore, we observe that different tasks require varying amounts of downstream data to reduce the evaluation loss below 0.30, which correlates with the length and difficulty of the tasks.

The results for Q2 are illuminated in \autoref{fig:scaling_law_v2}. We also explore the relationship between the evaluation loss during post-training on non-trajectory vision-language tasks and task success rate in downstream tasks. We use base models from different stages of post-training (with different eval loss on post-training datasets), fine-tuning them with the same downstream trajectory dataset. The baseline represents post-training using imitation learning on cross-task trajectories. We find that, for nearly all tasks, the success rate in downstream tasks correlates linearly with evaluation loss in post-training, with the lowest loss yielding the best results. Notably, models post-trained with knowledge-based tasks exhibit the best downstream performance for a given evaluation loss. Models enhanced with spatial grounding show the lowest evaluation loss and the highest task success rates.
These findings demonstrate scaling up off-trajectory vision language datasets directly enhances downstream task performance, which has been overlooked in previous VLA works~\citep{openVLA,rt-2}. 

\section{Related Works}
\label{Preliminary}

\subsection{Visual-Language-Action Models}
Imitation learning (IL) involves learning by mimicking expert interactions with the environment, with the primary challenge being the collection of high-quality expert demonstration datasets. Numerous studies have sought to enhance traditional IL approaches~\citep{diffusion_policy,octo,rt-1}. A promising direction is the use of Visual-Language-Action (VLA) models~\citep{rt-2,openVLA,tracevla,hiRT,dexgraspvla}, which adopt end-to-end imitation learning by fine-tuning VLMs. OpenVLA~\citep{openVLA} has demonstrated the importance of selecting a capable VLM backbone, a conclusion further reinforced by RoboVLM~\citep{roboVLA}. Similarly, \citet{rt-2} highlighted that co-training with web-scale vision-language data significantly improves the generalization of VLA models. While previous works primarily focused on optimizing the selection of VLMs, several recent studies have begun to pay attention to the comprehension capabilities of VLA models~\citep{upvla,objectvla,chatvla,combat_vla}. However, few have explicitly focused on enhancing the VLM backbone itself through visual-language post-training. Our work addresses this gap by proposing targeted visual-language post-training methods to enrich the capabilities of VLMs, thereby improving their performance on downstream VLA tasks.

\subsection{VLM-based Agents in Minecraft}\label{app:vlm_agents_minecraft}
Existing Minecraft agents based on VLMs typically adopt hierarchical architectures~\citep{deps,minedreamer,minedojo,deng2025open,zhao2024optimizing,cai2023open}. These methods leverage a VLM’s world knowledge for planning via zero-shot or few-shot in-context learning, without modifying the VLM parameters during agent optimization~\citep{deps,jarvis1,voyager,li2024optimus}.
STEVE-EYE~\citep{zheng2023steve} fine-tuned Llama language models~\citep{llama} using internet text data, achieving improved planning over zero-shot prompting. MineDreamer~\citep{minedreamer} employs the instruction-following capability of VLMs to predict future visual observations and generate actions based on STEVE-1~\citep{steve1}. OmniJARVIS~\citep{omnijarvis} uses a behavior tokenizer~\citep{groot,groot2} to model human trajectories in Minecraft with pretrained VLMs.
While these approaches optimize VLMs, they still rely on additional policies for action grounding. In contrast, we propose a VLA-based agent model that generates actions directly from textual instructions and visual inputs, eliminating the need for extra grounding policies.

\section{Conclusions}
We present \method, a novel training framework for visual-language-action models that leverages vision-language post-training to enhance decision-making capabilities in dynamic environments. Our experiments demonstrate that post-training on non-trajectory tasks significantly enhances foundation models’ ability to understand complex environments, resulting in substantial improvements in downstream imitation learning on trajectory data. The effectiveness of this model is validated across multiple VLM architectures, providing strong evidence of its broad applicability and potential for visual-language-action model training, as exemplified by our state-of-the-art model, \model.

%% file: tables/atom_tasks.tex
\begin{table*}[t]
\centering
\resizebox{0.99\linewidth}{!}{
\renewcommand\arraystretch{1.1}
\begin{tabular}{@{}lccccccccccccccccc@{}}
\toprule
\multirow{2}{*}{Model} & \multirow{2}{*}{Model Size} & \multicolumn{3}{c}{\textbf{\texttt{Mine Blocks}}} &  & \multicolumn{3}{c}{\textbf{\texttt{Kill Entities}}} &  & \multicolumn{4}{c}{\textbf{\texttt{Craft Items}}} &  & \multicolumn{3}{c}{\textbf{\texttt{Smelt Items}}} \\ \cmidrule(lr){3-5} \cmidrule(lr){7-9} \cmidrule(lr){11-14} \cmidrule(l){16-18} 
 &  & \includegraphics[scale=0.045,valign=c]{figures/avatar/iron_ore.png} & \includegraphics[scale=0.035,valign=c]{figures/avatar/obsidian.png} & Avg. &  & \includegraphics[scale=0.05,valign=c]{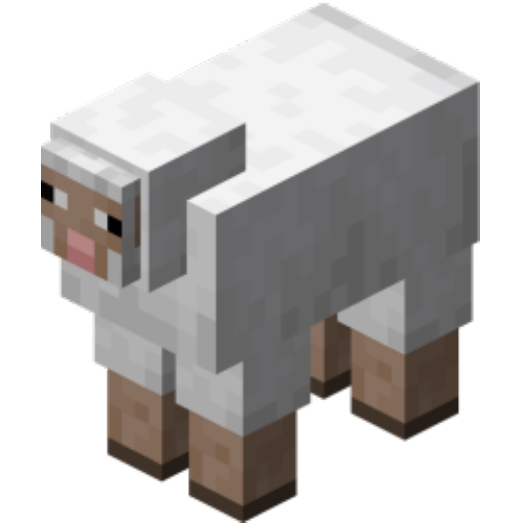} & \includegraphics[scale=0.055,valign=c]{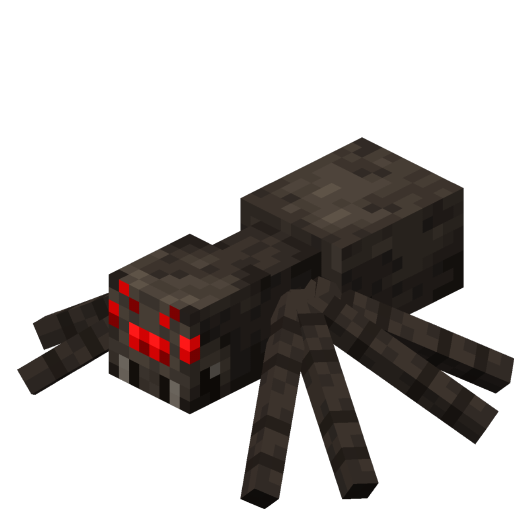} & Avg. &  & \includegraphics[scale=0.08,valign=c]{figures/avatar/crafting_table.png} & \includegraphics[scale=0.07,valign=c]{figures/avatar/diamond_sword.png} & \includegraphics[scale=0.07,valign=c]{figures/avatar/ladder.png} & Avg. &  & \includegraphics[scale=0.04,valign=c]{figures/avatar/cooked_beef.png} & \includegraphics[scale=0.04,valign=c]{figures/avatar/iron_ingot1.jpg} & Avg. \\ \midrule
VPT-BC~\citep{vpt} & 248M & 0.15 & 0.38 & 0.33 &  & \color{blue}{{0.55}} & 0.35 & 0.44 &  & 0.30 & \color{blue}{{0.50}} & 0.45 & 0.41 &  & 0.10 & 0.00 & 0.05 \\
VPT-RL~\citep{vpt} & 248M & 0.05 & 0.35 & 0.25 &  & 0.35 & 0.25 & 0.28 &  & \color{blue}{{0.50}} & 0.30 & 0.62 & 0.55 &  & 0.05 & 0.35 & 0.20 \\
STEVE-1~\citep{steve1} & 248M & 0.20 & 0.35 & 0.54 &  & 0.30 & \color{blue}{{0.75}} & 0.38 &  & 0.45 & 0.20 & \color{blue}{{0.70}} & \color{blue}{{0.57}} &  & 0.25 & \color{blue}{{0.40}} & \color{blue}{{0.33}} \\
GROOT~\citep{groot} & 248M & \color{blue}{{0.56}} & \color{blue}{{0.40}} & \color{blue}{{0.67}} &  & 0.50 & 0.50 & \color{blue}{{0.52}} &  & 0.45 & 0.35 & 0.25 & 0.40 &  & \color{blue}{{0.35}} & 0.25 & 0.30 \\ 
MineDreamer~\citep{minedreamer} & 7B & 0.25 & \color{blue}{{0.40}} & 0.55 &  & 0.30 & 0.70 & 0.39 &  & \color{blue}{{0.50}} & 0.25 & 0.30 & 0.42 &  & 0.30 & 0.30 & 0.30 \\ \midrule
Qwen2-VL~(raw) & 7B & 0.77 & 0.60 & 0.79 &  & 0.93 & 0.80 & 0.84 &  & 0.83 & 0.53 & 0.40 & 0.60 &  & 0.03 & 0.10 & 0.07 \\
Qwen2-VL~(IL) & 7B & 0.70 & 0.73 &  0.75 & & 0.97 & 0.83 & 0.86 & & 0.73 & 0.67 & 0.50 &  0.65 & & 0.17 & 0.37 & 0.29 \\
\model-Qwen2 & 7B & \color{red}{\textbf{0.80}} & \color{red}{\textbf{0.95}} & \color{red}{\textbf{0.88}} &  & \color{red}{\textbf{0.97}} & \color{red}{\textbf{0.93}} & \color{red}{\textbf{0.95}} &  & \color{red}{\textbf{0.87}} & \color{red}{\textbf{0.83}} & \color{red}{\textbf{0.63}} & \color{red}{\textbf{0.77}} &  & \color{red}{\textbf{0.77}} & \color{red}{\textbf{0.70}} & \color{red}{\textbf{0.70}} \\ \bottomrule
\end{tabular}}
\caption{
Evaluation results of different policies on Minecraft tasks, 
Each group includes multiple tasks (at least 5), and the Avg. column reports the average success rate within each group. Qwen2-VL, Qwen2-VL~(IL) and \model-Qwen2-VL represent the training on the original qwen checkpoint, post-training on only large-scale imitation learning trajectories, and post-trained on VLP intermediate model. Qwen2-VL~(\method) achieves the highest success rates across all task groups. 
}
\label{tab:atom_tasks}
\end{table*}

%% file: tables/paradigm_ablation.tex
\begin{table}[t]
\centering

\resizebox{0.98\linewidth}{!}{
\begin{tabular}{>{\raggedright\arraybackslash}p{4.5cm} ccccccc} 
\toprule
\textbf{Model} & \includegraphics[scale=0.08,valign=c]{figures/avatar/diamond_sword.png} & \includegraphics[scale=0.07,valign=c]{figures/avatar/ladder.png} & \includegraphics[scale=0.04,valign=c]{figures/avatar/cooked_beef.png} & 
\includegraphics[scale=0.04,valign=c]{figures/avatar/iron_ingot1.jpg}  \\
\midrule
Qwen2-VL(raw) & 0.53 & 0.40 & 0.03 & 0.10 \\
Qwen2-VL-7B (one-stage) & 0.10 & 0.40 & 0.07 & 0.13 \\
\method-Qwen2-VL & 0.83 & 0.63 & 0.77 & 0.70 \\
\bottomrule
\end{tabular}}
\caption{\textbf{Ablation results for different training paradigms.} We use Qwen2-VL (raw) as the baseline and compare it against Qwen2-VL-7B (one-stage) and \model-Qwen2-VL on four Minecraft tasks: "craft diamond sword", "craft ladder", "cook beef", and "smelt iron ingot".}
\label{tab:paradigm_ablation}
\end{table}

%% file: appendix.tex
\section{Observation and Action Space}\label{app:action_space}
We rely solely on visual images for observation, without any symbolic information, similar to VPT~\citep{vpt}. Notably, we increase the image resolution to 644×364, providing significantly more visual detail than VPT (128×128), which helps \model better perceive the environment.

To closely mimic human behavior, our action space includes all possible player actions except for arbitrary text input. We allocate reserved tokens outside the original VLM vocabulary to represent keypress and click actions. For mouse movements, we follow VPT~\citep{vpt}, using $\mu$-law encoding to discretize X and Y axes separately into 21 bins (42 tokens total), each mapped to a reserved token. Although Qwen2-VL~\citep{qwen2vl} does not explicitly support reserved tokens like Llama3~\citep{llama3.2}, we can add special tokens by expanding the vocabulary, since its embedding table is underutilized. \autoref{tab:minecraft_action_space} shows the action space \model uses. During inference, the model predicts actions token by token: key/button actions first, followed by camera movements (Pitch and Yaw).

\begin{table}[htbp] 
\centering 
\caption{The action space we use in Minecraft.} 
\label{tab:minecraft_action_space} 
\begin{tabularx}{\linewidth}{r l l X} 
\toprule
\textbf{Index} & \textbf{Action} & \textbf{Human Action} & \textbf{Description} \\
\midrule
1 & Forward & key W & Move forward. \\
2 & Back & key S & Move backward. \\
3 & Left & key A & Strafe left. \\
4 & Right & key D & Strafe right. \\
5 & Jump & key Space & Jump. When swimming, keeps the player afloat. \\
6 & Sneak & key left Shift & Slowly move in the current direction of movement. \\
7 & Sprint & key left Ctrl & Move quickly in the direction of current motion. \\
8 & Attack & left Button & Destroy blocks (hold down); Attack entity (click once). \\
9 & Use & right Button & Interact with the block. \\
10 & hotbar:[1-9] & keys 1 - 9 & Selects the appropriate hotbar item. \\
11 & Inventory & key E & Open/Close the inventory. \\
12 & Yaw & move Mouse X & Turning; aiming; camera movement. \\
13 & Pitch & move Mouse Y & Turning; aiming; camera movement.  \\
\bottomrule
\end{tabularx}
\end{table}

\section{Training Configurations}\label{app:training_configs}

The training configurations for both Visual-Language Post-Training and Action Post-Training are largely 
consistent. All experiment were conducted on NVIDIA A800-SXM4-80GB GPUs, utilizing CUDA version 12.1 and Hugging Face Transformers version 4.47.0. 
Both training stages utilized the AdamW optimizer with $\beta_1 = 0.9$, $\beta_2 = 0.95$, weight decay were set to 0, and $\epsilon = 1\times10^{-8}$. A cosine learning rate schedule was adopted with the learning rate of $5\times10^{-6}$ and a warmup of 200 steps. The training used \texttt{bfloat16} precision, a maximum gradient norm of 1.0, and a fixed random seed of 42. To accelerate training, DeepSpeed with ZeRO-1~\citep{deepspeed}optimization was employed. For \textbf{Visual-Language Post-Training}, the maximum token length was set to 3584, and we set 
a batch size per device of 2 and a gradient accumulation of 4
. For \textbf{Action Post-Training}, the maximum token length was set to 512, which allowed a batch size per device of 8 and a gradient accumulation of 1 step per update. Ensuring that the total batch size remained 256. Both stages were trained using 32 A800 GPUs, with the Visual-Language Post-Training phase running for 128 GPU hours and the Action Post-Training phase running for 512 GPU hours.

To enhance generalization, distinct data augmentation strategies were adopted for different training phases. In the \textbf{Visual-Language Post-Training} phase, modifications included adjustments to hue, saturation, brightness, contrast, as well as random translation, rotation, slight scaling variations, shearing, and occasional flipping. These adjustments extended to bounding box and pointing annotations, with necessary masking of instruction-following prompts. In contrast, the \textbf{Action Post-Training} phase focused on adjusting hue, saturation, brightness, contrast, and translation, applied only on images.

\section{Details of Inference}
During inference, \model is expected to output actions in a format consistent with the gameplay dataset, as illustrated below:

\begin{prompt}[title={Example of \model Interaction for One Turn of Iteration}, label=] 
\begin{center}
\begin{minipage}{\dimexpr0.95\linewidth}
\vspace{5pt}

\textbf{Instruction}: Craft a bread so I can use it.

Arrange the materials in the crafting grid according to the following pattern:

 wheat | wheat | wheat 

 wheat | wheat | wheat 

 and get 1 bread. 

\textbf{Observation}: 

\vspace{5pt}
\begin{minipage}{\linewidth}
  \includegraphics[width=0.6\linewidth]{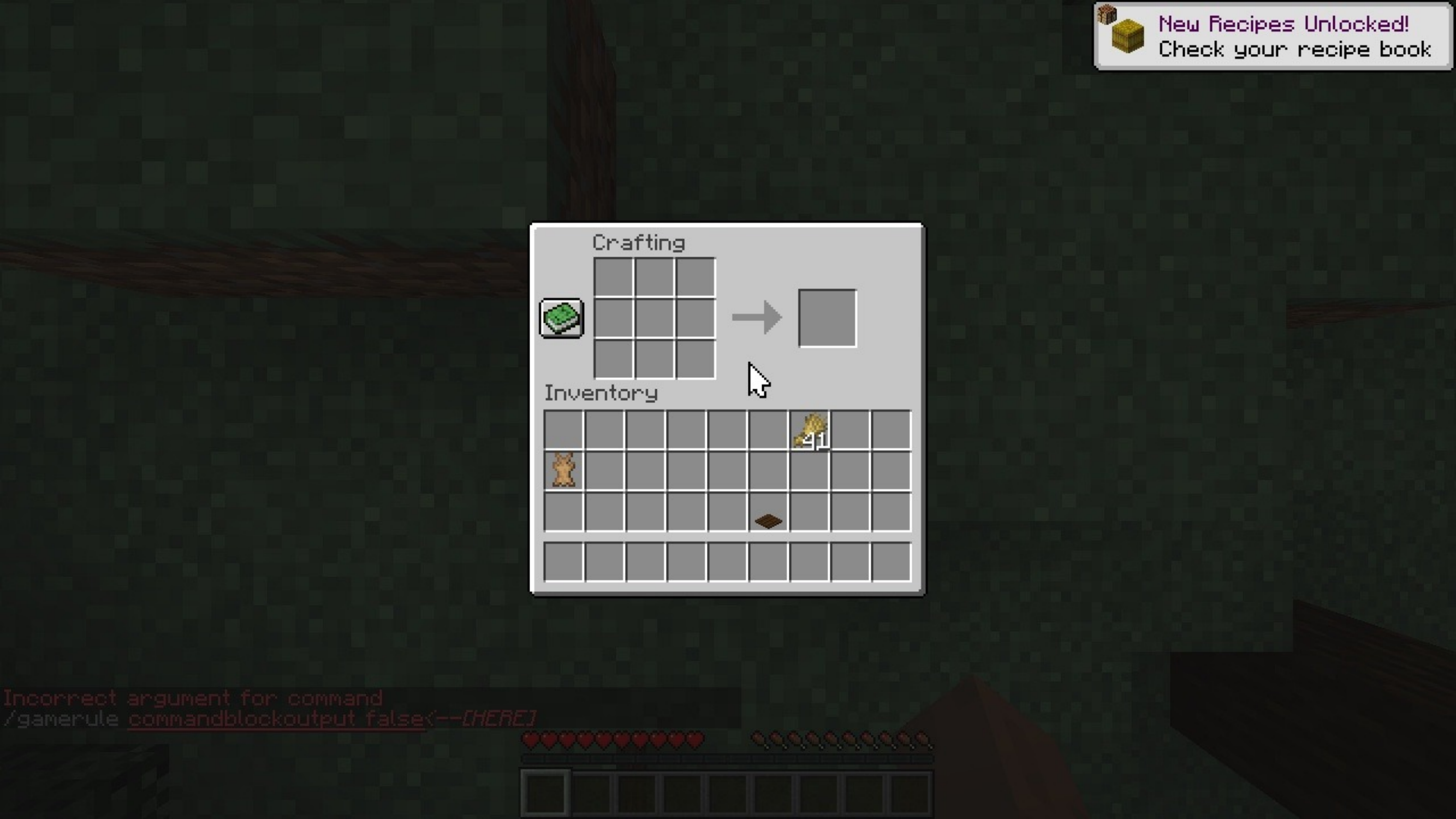}
  \label{fig:app vlpt inference} 
\end{minipage}
\vspace{5pt}

\textbf{Action}: 

\textless|action\_begin|\textgreater \textless|cam\_w\_13|\textgreater \textless|cam\_h\_5|\textgreater \textless|action\_end|\textgreater

\textless|action\_begin|\textgreater\textless|use|\textgreater \textless|cam\_w\_3|\textgreater \textless|cam\_h\_2|\textgreater \textless|action\_end|\textgreater

\vspace{3pt}
\end{minipage}
\end{center}
\end{prompt}

\subsection*{Action Chunking Strategy}
To optimize both inference speed and predictive performance, we designed an adaptive action chunking strategy that varies the number of predicted actions per observation across different operational phases. The strategy begins with a chunk size of one action per observation during Action Post-Training to ensure precise policy alignment. In the subsequent fine-tuning stages, the chunk size is increased to three to accelerate learning. Finally, during inference, the model generates two actions per prediction step, balancing stability and efficiency.

To validate this multi-phase chunking approach, we conducted experiments on the MCU~\citep{mcu} tasks using Qwen2-VL-2B~\citep{qwen2vl} as the base model. The results are presented in ~\autoref{tab:action_chunk_ablation}. This ablation study suggests that employing longer action chunks during advanced training stages and a moderate chunk size for inference can improve both success rates and FPS.

\begin{table}[htbp]
\centering
\caption{Ablation study on action chunk sizes. "APT Chunks" denotes Action Post-Training chunks, and "FT Chunks" denotes Subsequent Fine-tuning chunks.}
\label{tab:action_chunk_ablation}
\begin{tabular}{ccccc}
\toprule
APT Chunks & FT Chunks & Inference Chunks & FPS & Success Rate \\
\midrule
1 & 1 & 1 & 8  & 0.47 \\
1 & 3 & 1 & 8  & 0.53 \\
1 & 3 & 2 & 15 & \textbf{0.67} \\
1 & 3 & 3 & \textbf{21} & 0.20 \\
3 & 3 & 1 & 8  & 0.03 \\
\bottomrule
\end{tabular}
\end{table}

As shown in \autoref{tab:action_chunk_ablation}, the chunking strategy of using one action per observation during Action Post-Training, three during Further Fine-tuning, and two during Inference (1 $\rightarrow$ 3 $\rightarrow$ 2) achieved the highest success rate (0.67) and a solid frame rate (15 FPS) in this specific ablation. We did not explore chunk sizes larger than three, as the Minecraft environment changes rapidly with each discrete action. Predicting multiple future actions in such a dynamic setting poses a significant challenge.

It is also important to clarify the choice of inference chunk size in our main results. To ensure a rigorous and fair comparison with baseline models such as Qwen2-VL (raw) and Qwen2-VL (IL)—which both operate with a single predicted action—we configured \model to use an inference chunk size of one in our primary experiments. While this setting facilitates direct comparison, our ablation results~\ref{tab:action_chunk_ablation} suggest that a chunk size of two at inference may further improve both efficiency and task success under alternative evaluation criteria.

\subsection*{Infrastructure for Efficient Inference}
To facilitate efficient inference, \model was served using the vLLM~\citep{vllm} in \texttt{bfloat16} precision. This setup was typically deployed on four A800 GPUs. Compared to a standard Hugging Face Transformers implementation, vLLM increased the inference throughput of \model by up to 1.7 times, without introducing any significant degradation in task performance.

\subsection*{Achieved Inference Performance}
The final inference speed reflects both the action chunking strategy and infrastructure optimizations. In our main results, which use single-action inference for fair comparison, \model-Qwen2-VL-7B runs at 5.5 FPS, while \model-Qwen2-VL-2B achieves 8 FPS—consistent with the results in rows 1 and 2 of \autoref{tab:action_chunk_ablation}. Notably, \model-Qwen2-VL-2B can reach 15 FPS with a chunk size of two, highlighting a trade-off we made in favor of direct benchmarking with baseline models.

\section{Datasets}\label{app:training_datasets}

In this section, we will introduce the collection methods and specific forms of the visual language post-training datasets we used, including the world knowledge dataset, visual language alignment dataset, and visual grounding dataset.

\subsection{World Knowledge Dataset}

We sourced content from Wikipedia and the DigMinecraft website, and generating 202K question answering entries using the GPT-3.5-turbo model. The dataset have been adjusted to a uniform format. To increase training efficiency, we randomly merged the original one-question-one-answer data to organize multi-turn dialogue dataset with lengths of approximately 1500-2000 tokens each. Below is the specific format of the world knowledge dataset.

\begin{prompt}[title={Example of World Knowledge Dataset}, label=World Knowledge Post Training Datasets]

\begin{center}
\begin{minipage}{\dimexpr0.95\linewidth}
\vspace{5pt}

\textbf{Question}: "What are the different items that can be made using dark oak planks in Minecraft?"

\textbf{Answer}: "In Minecraft, you can use dark oak planks to craft various items and blocks. Here are some of the items and blocks you can make with dark oak planks:

1. Dark Oak Wood Slabs: Used for flooring, roofs, and other building elements.

2. Dark Oak Wood Stairs: Ideal for creating inclined pathways or staircases.

3. Dark Oak Wood Fence: Useful for creating boundaries or enclosures around structures.

4. Dark Oak Wood Fence Gate: Allows for openings in your fences for easy access.

5. Dark Oak Wood Door: Adds entrances to your buildings.

6. Dark Oak Wood Trapdoor: Can serve as hidden entrances or functional hatches.

7. Dark Oak Wood Button: Can be placed on surfaces and used to activate various redstone mechanisms.

8. Dark Oak Wood Pressure Plate: Activates redstone contraptions when stepped on.

9. Dark Oak Wood Sign: Useful for displaying text and providing information in your world.

10. Dark Oak Wood Bed: Used to craft the frame of a bed, combined with wool blocks for a complete bed.

11. Dark Oak Wood Boat: Useful for water transportation.

12. Dark Oak Wood Chest: Can be used to craft storage chests.

13. Dark Oak Wood Bookshelf: Decorative and enhances enchantment tables.

14. Petrified Oak Slab: Made from dark oak logs submerged in water and then crafted into planks and slabs. These items enhance functionality and aesthetics in your Minecraft builds."

\rule{\linewidth}{0.4pt}

\vspace{4pt}

\textbf{Question}: "How can players build and use automatic farms in Minecraft, and what are some efficient designs?"

\textbf{Answer}: "In Minecraft, automatic farms are structures designed to automatically harvest resources without requiring constant player intervention. These farms can be extremely useful for gathering various resources efficiently. Here, I'll provide a general overview of how to build and use automatic farms, along with examples of efficient designs for different types of farms:

1. Crop Farms:
Automated crop farms are used to grow and harvest crops like wheat, carrots, potatoes, and more.
Efficient design:
- Create a water-based system to hydrate the farmland. A water source in the middle with canals leading to the farmland is a common approach.
- Use a dispenser to plant seeds or crops onto the farmland.
- Attach a redstone clock to the dispenser to periodically activate it.
- Collect the harvested items using hoppers and chests.

2. Mob Farms:
These farms are used to automatically collect drops from hostile mobs, such as zombies, skeletons, and creepers.
Efficient design:
- Build a dark room where mobs will spawn naturally. Use trapdoors, pressure plates, or a combination to restrict mob movement.
- Use water currents to funnel mobs into a kill chamber.
- Create a killing mechanism using either lava blades, suffocation, or other methods.
- Use water currents again to collect drops and funnel them into hoppers for collection.

3. Animal Farms:
Automated animal farms can help you gather resources from passive mobs like cows, pigs, and chickens.
Efficient design:
- Create separate pens for each type of animal.
- Use a breeding mechanism to maintain a stable population. Dispensers with food (like wheat for cows) can automatically breed animals.
- Include a killing mechanism for excess animals, providing a steady supply of drops.
- Use water currents and hoppers to collect drops into chests.

4. Tree Farms:
Automated tree farms allow you to grow and harvest trees without manual effort.
Efficient design:
- Plant saplings in rows using dispensers and bone meal.
- Use a mechanism to cut down the trees. This can be achieved using flying machines or TNT-based contraptions.
- Use water currents to collect the dropped wood and saplings into hoppers and chests.

5. Fishing Farms:
Automated fishing farms simulate fishing to gather loot and experience points.
Efficient design:
- Build a platform with water sources, a fishing rod dispenser, and a mechanism to reel in the fishing rod.
- Create a hopper system to collect items caught from the fishing process.


Remember that the efficiency of these designs can vary depending on your Minecraft version and specific game settings. Additionally, always consider ethical and environmental factors when designing and using automated farms in your gameplay."
\vspace{3pt}
\end{minipage}
\end{center}

\end{prompt}

\subsection{Vision Language Alignment Dataset}

For the construction of Vision Language Alignment Dataset, we employed a multi-step approach to gather and process data. Initially, we sourced images from a variety of online platforms, including YouTube gameplay videos, contractor-provided screenshots \citep{vpt}, and other internet resources. We carefully selected 35,000 keyframes from these sources based on criteria such as brightness and visual complexity to ensure the quality of the dataset. Subsequently, we utilized advanced Vision-Language Models—such as GPT-4o \citep{chatgpt}, Claude 3.5 Sonnet \citep{claude}, and Molmo \citep{molmo}—to generate contextual image question-answer pairs, critical in creating a rich, semantically detailed dataset that bridges visual content and linguistic annotations. To enhance the reliability of the generated data, we implemented a robust query-validation pipeline. This pipeline employed Llama-3.1-70B \citep{llama-3.1} to systematically filter out ambiguous questions and validate the accuracy of the generated answers. Additionally, it included a validation step to ensure that the question-answer pairs were relevant to the associated images. To further diversify our dataset, we specifically allocated nearly half of the vision-question-answer pairs for caption generation, employing the advanced capabilities of GPT-4o. Through the implementation of our comprehensive pipeline, we successfully generated a Vision Language Alignment Dataset comprising 15,000 captions and 20,000 visual question answering dataset.

\begin{prompt}[title={Example of Vision Language Alignment Datasets}, label=Vision Language Alignment Datasets]

\vspace{5pt}

\begin{minipage}{\linewidth}
  \centering
  \includegraphics[width=0.6\linewidth]{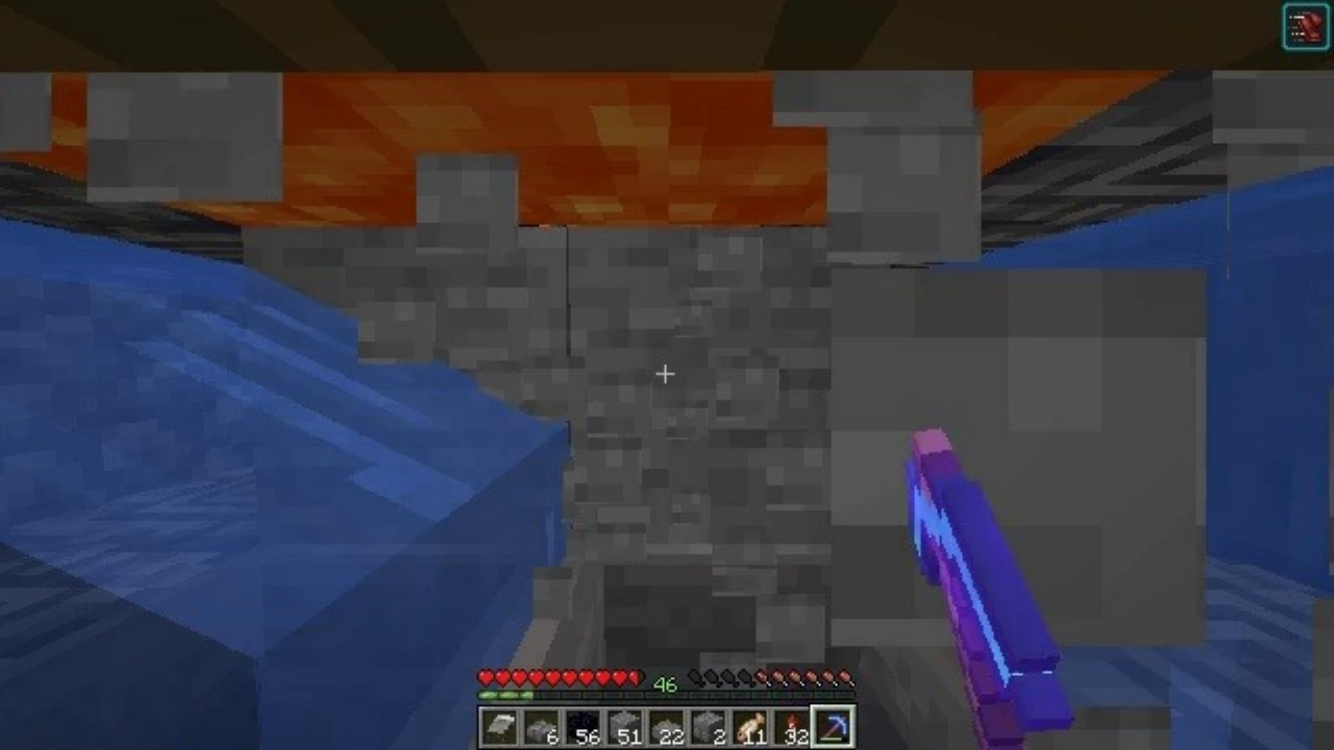}
  \label{fig:app vl}
\end{minipage}
\vspace{15pt}


\begin{center}
\begin{minipage}{\dimexpr0.95\linewidth} 

\textbf{Question}: "What tool is the player holding?"

\vspace{5pt}

\textbf{Answer}: "The player is holding a pickaxe with an enchantment glow, likely a diamond pickaxe due to its color."






\rule{\linewidth}{0.4pt}

\vspace{4pt}

\textbf{Question}: "What type of block is being mined by the player?"

\vspace{5pt}

\textbf{Answer}: "The player is mining a stone block."

\rule{\linewidth}{0.4pt}

\vspace{4pt}

\textbf{Question}: "What color is the liquid at the top of the image?"

\vspace{5pt}

\textbf{Answer}: "The liquid at the top of the image is orange, indicating it is lava."

\rule{\linewidth}{0.4pt}

\vspace{4pt}

\textbf{Question}: "How many inventory slots does the player have in total (visible on the hotbar)?"

\vspace{5pt}

\textbf{Answer}: "The player has 9 inventory slots visible on the hotbar."

\end{minipage} 
\end{center} 

\vspace{5pt}

\end{prompt}


\subsection{Visual Grounding dataset}

For Visual Grounding Dataset, we adopted different collection methods for 3D environments and GUI interactions within Minecraft. 

In 3D environment, we sample images from the contractor data. We initially identify objects that are interacting with the player at a specific moment in the video. Then implementing Backward Trajectory Relabeling~\citep{rocket1} by sam2~\citep{sam2} to accurately trace the positions of objects up to $t$ frames backward, which tend to be in more random positions.

For 2D GUI interactions, we directly utilized inventory messages provided by Minecraft in rollouts to determine the positions of items within the inventory. Given that each slot in the inventory has a fixed position, we can directly ascertain the points of items. To enhance the randomness of the data distribution, we performed cropping on the data to make the point distribution more uniformly distributed.

In total, we have collected 404K point data from these efforts, mapping real-time entity and object positions during scripted tasks effectively.

\begin{prompt}[title={Example of Vision Grounding Datasets}, label=Vision Language Grounding Datasets]

\begin{center}
\begin{minipage}{\dimexpr0.95\linewidth}
\vspace{5pt}

\textbf{Question}: "Show the location of torch in the provided image by outlining it with a bbox."
\vspace{5pt}

\begin{minipage}{\linewidth}

  \includegraphics[width=0.6\linewidth]{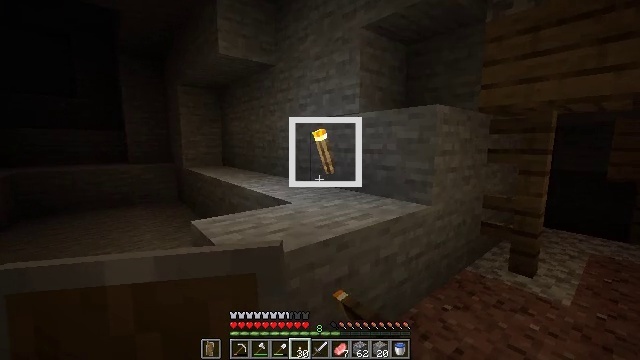}
  \label{fig:app-ground1}
\end{minipage}
\vspace{7pt}

\textbf{Answer}: "\textless|object\_ref\_start|\textgreater torch\textless|object\_ref\_end|\textgreater\textless|bbox\_start|\textgreater(453,333),(563,528)\textless|bbox\_end|\textgreater"

\vspace{5pt}

\rule{\linewidth}{0.4pt}

\vspace{5pt}

\textbf{Question}: "Illustrate the precise position of the iron boots in this image, returning the location in points."
\vspace{5pt}

\begin{minipage}{\linewidth}

    \includegraphics[width=0.6\linewidth]{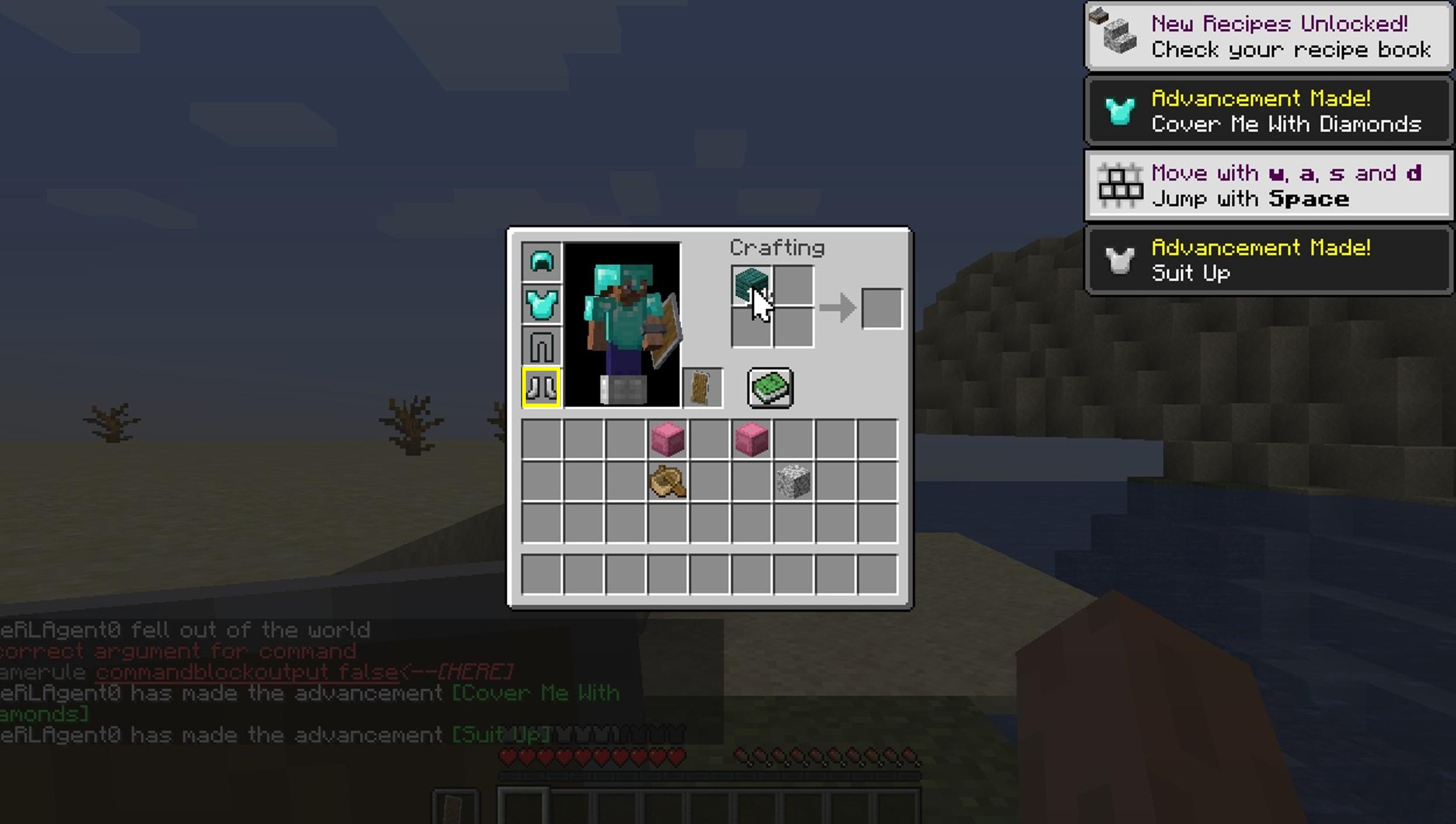}
    \label{fig:app-ground2}
\end{minipage}
\vspace{7pt}

\textbf{Answer}: "\textless|object\_ref\_start|\textgreater iron boots\textless|object\_ref\_end|\textgreater\textless|point\_start|\textgreater(356,446),(386,494)\textless|point\_end|\textgreater"

\end{minipage}
\end{center}

\end{prompt}


\section{Benchmarks}\label{app:benchmarks}

In this section, we will introduce our benchmark to test the capability of the Visual Language Models. We also divide the benchmarks into three sections: world knowledge evaluation, visual recognition evaluation, and spatial grounding evaluation. For the first two datasets, we wrote most of them ourselves, while others were obtained from freely available examples on the Internet. We organized them into open question and answer formats. For the spatial grounding evaluation tasks, we adopted the same method to collect data as for the Visual Grounding Dataset~\ref{app:training_datasets}, ensuring that the test set does not overlap with the training dataset. Below is a detailed introduction:

\subsection{World Knowledge Evaluation}

For World Knowledge Evaluation, we divided it into eight subcategories: craft, plant, mobs, kill, enchant, potion, biomes, and architecture, with a total of 47 test questions. For each entry, we provided the correct answer. A detailed description is shown in \autoref{table:knowledge-evaluation}.

\begin{table}[ht]
\centering
\caption{Summary of Minecraft knowledge questions and answers evaluation.}
\label{table:knowledge-evaluation}
{\linespread{1.0}\selectfont
\resizebox{0.98\linewidth}{!}{
\begin{tabular}{c c p{6.2cm} p{6.2cm}}
\toprule
\textbf{Category} & \textbf{Quantity} & \textbf{Example Question} & \textbf{Example Answer} \\
\midrule
Craft        & 9  & What materials are needed to craft a jukebox in Minecraft? & 8 Planks and 1 Diamond. \\
Plant        & 5  & What is the maximum height sugarcane can reach in Minecraft? & 3. \\
Mobs         & 12 & What happens when a creeper gets struck by lightning in Minecraft? & A creeper becomes a charged creeper. \\
Kill         & 3  & Can shield block crossbow attack in Minecraft? & No. \\
Enchant      & 3  & What happens if the player puts on an item with the Curse of Binding enchantment in Minecraft? & It cannot be removed until the item breaks or the player dies. \\
Potion       & 4  & What materials are needed to craft a potion of poison in Minecraft? & Water bottle, Nether wart, blaze powder, spider eye. \\
Biomes       & 4  & Which biome is the only place you can find blue orchids in Minecraft? & Swamp. \\
Architecture & 7  & How many chests are typically found in a shipwreck in Minecraft? & 1 to 3. \\
\bottomrule
\end{tabular}
}
}
\end{table}

\begin{prompt}[title={Knowledge Evaluation}, label=Knowledge Evaluation]
\textbf{Question:} In Minecraft, what type of tool is best suited for digging Mycelium?

\textbf{GPT-4o:} A shovel is the best tool for digging Mycelium. [Explanation: In Minecraft, Mycelium is a block that can be quickly and efficiently broken using a shovel. Like other dirt-based blocks, using a shovel will speed up the process, making it much faster than breaking it by hand or with other tools. Additionally, enchantments like Efficiency on your shovel can further increase the speed at which you can dig Mycelium.]

\textbf{Qwen2-VL:} A Pickaxe. [Explanation: Mycelium is a special type of grass block in Minecraft that can only be found in The Nether, a dimension that can only be entered after the player has crafted the Nether Portal. To dig Mycelium, you will need to have a Pickaxe in your inventory. The Pickaxe is a tool that is used for digging, breaking, and mining blocks. It is the most efficient and effective tool for digging Mycelium because it has a higher efficiency for breaking blocks than other tools.]

\textbf{\model-Qwen2-VL:}A shovel is the best tool for digging Mycelium in Minecraft.
\end{prompt}

\subsection{Vision Understanding Evaluation}
For Vision Understanding Evaluation, we divided it into five subcategories: scene, object, info, inventory, and OCR, comprising a total of 43 test questions. For each entry, we provided an image along with the correct answer. A detailed description can be found in \autoref{table:vision-evaluation}. Specifically, the Scene category involves the game environment, including biomes, time, and weather; the Object category relates to objects within the player’s view; and the Info category includes aspects of the player’s experience, health, hunger, and hotbar status.

\begin{table}[htbp]
\centering

\caption{Summary of Vision Understanding Evaluation.}
\label{table:vision-evaluation}
\begin{tabular}{@{} >{\raggedright\arraybackslash}p{2cm} >{\centering\arraybackslash}p{1cm} >{\centering\arraybackslash}p{3cm} >{\raggedright\arraybackslash}p{3cm} >{\raggedright\arraybackslash}p{3cm} @{}}
\toprule
\textbf{Category} & \textbf{Quantity} &  \textbf{Image} & \textbf{Example Question} & \textbf{Example Answer} \\
\midrule
Scene & 11 &\multirow{1}{*}{\includegraphics[width=0.1\textwidth]{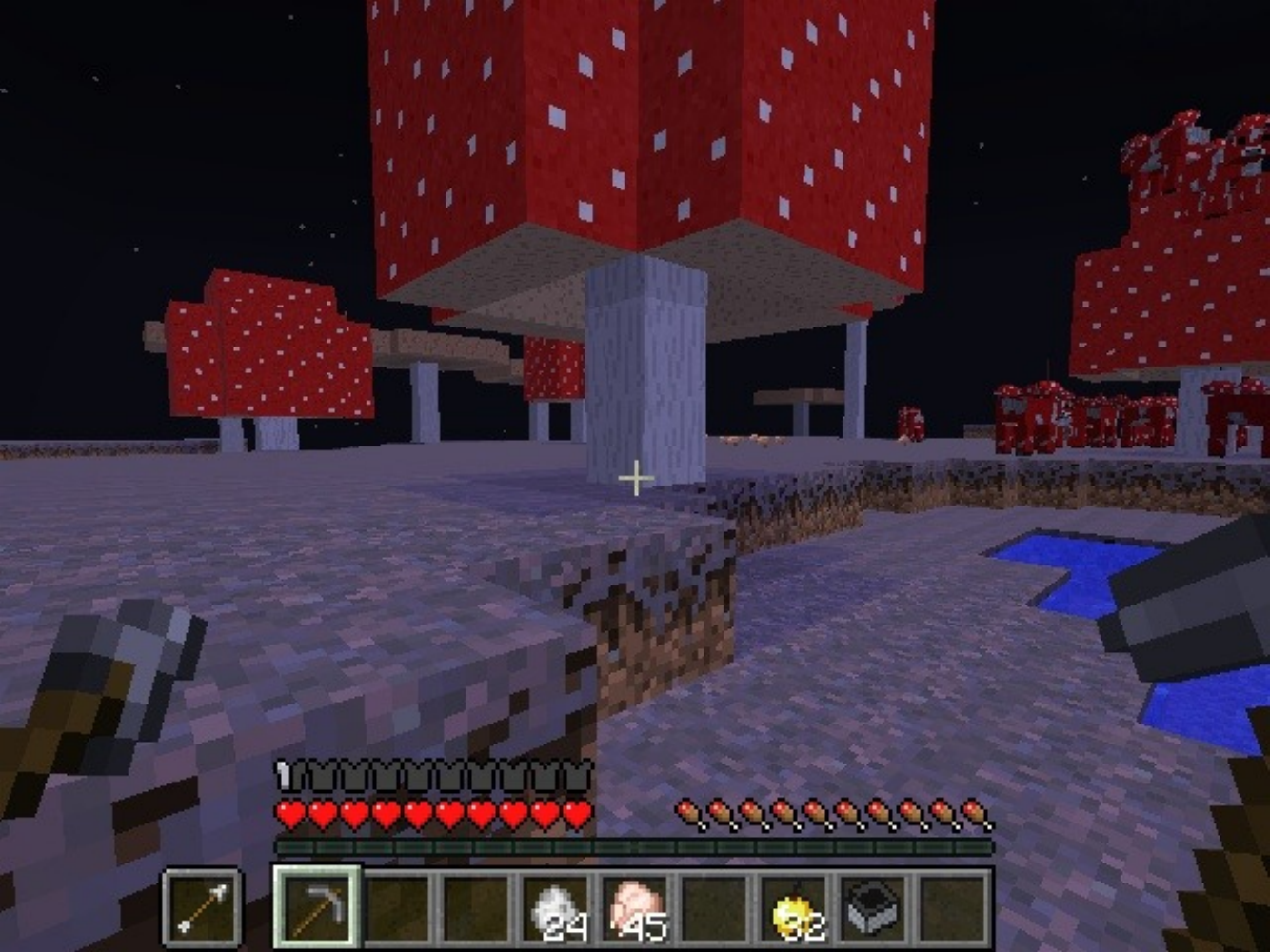}} & What biome you think is the player currently in? & Mushroom Fields biome. \\ \midrule
Object & 13 & \multirow{1}{*}{\includegraphics[width=0.1\textwidth]{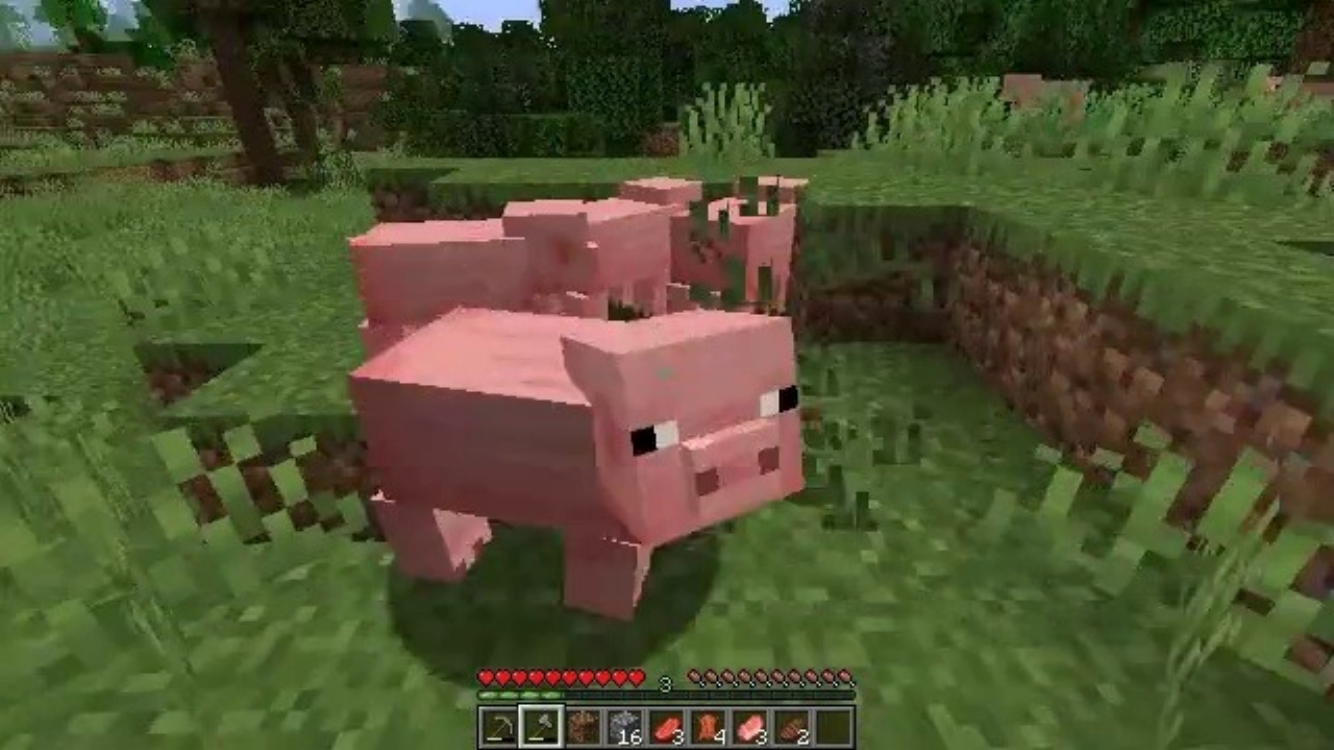}} & Are there any hostile mobs in the picture? & No, there are only 4 pigs in the picture.  \\ \midrule
Info & 7 & \multirow{1}{*}{\includegraphics[width=0.1\textwidth]{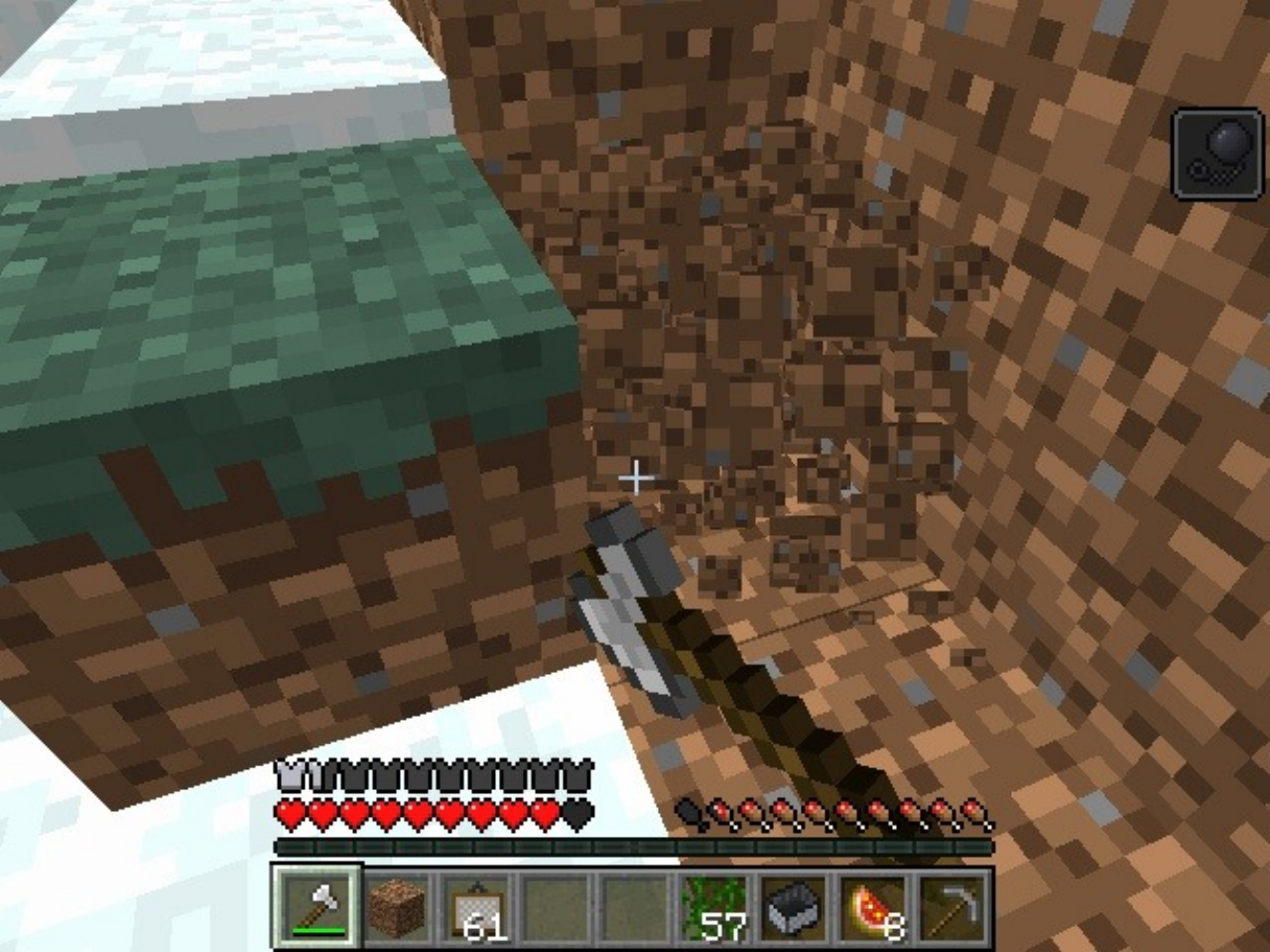}} & Is the player's hunger bar currently full? & No. \\ \midrule
Inventory & 6 &  \multirow{1}{*}{\includegraphics[width=0.1\textwidth]{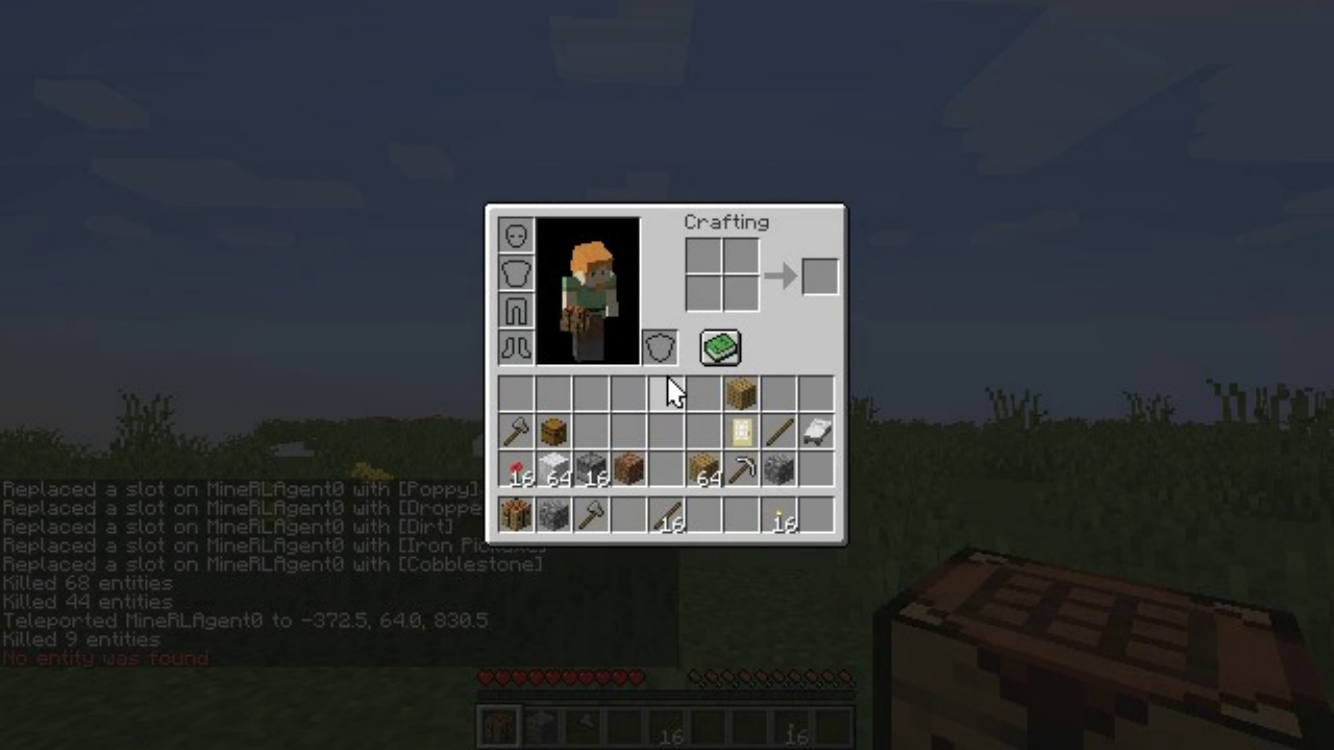}} & Is there any oak wood in the inventory? & Yes, there are oak wood planks in the inventory. \\ \midrule
OCR & 6  & \multirow{1}{*}{\includegraphics[width=0.1\textwidth]{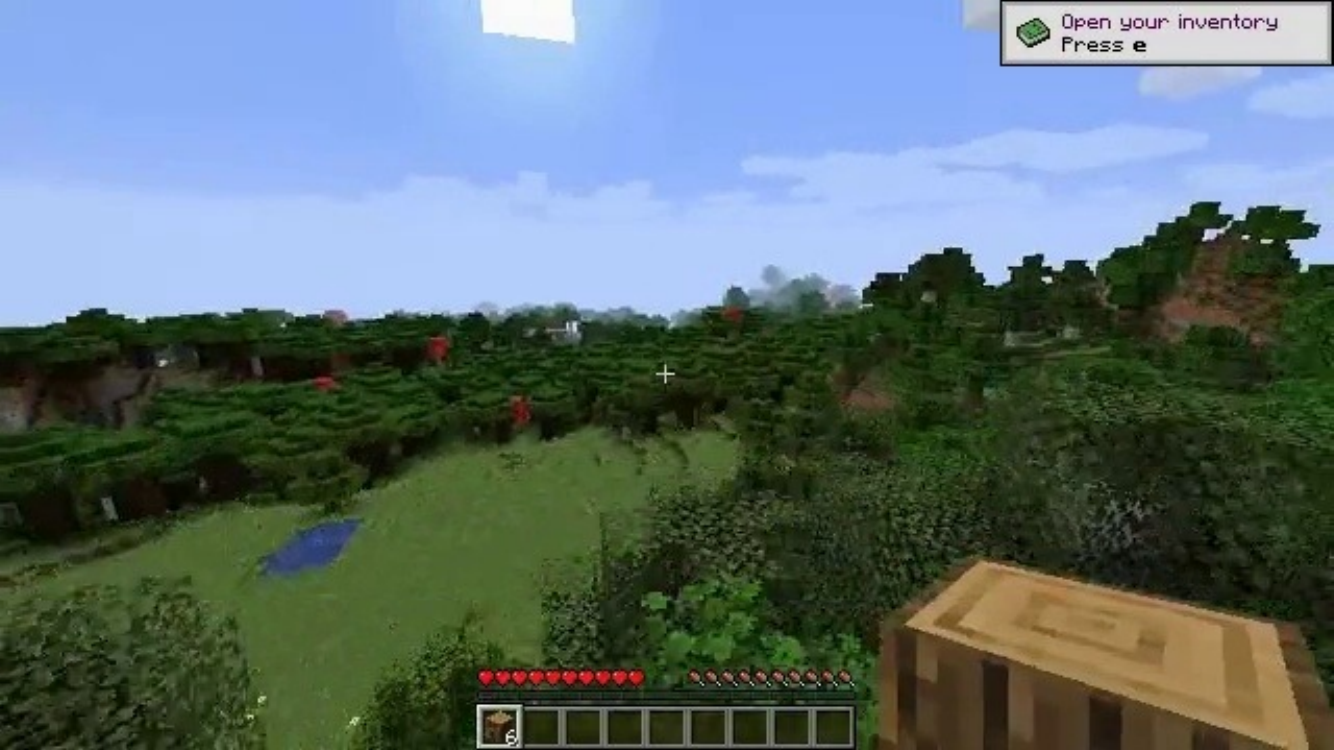}} & What instructions are visible on the screen? & Open your inventory Press e. \\
\bottomrule
\end{tabular}
\end{table}

\begin{prompt}[title={Vision Understanding Evaluation Examples}, label=Vision Understanding Evaluation]

\begin{center}
\begin{minipage}{\dimexpr0.95\linewidth}

\begin{minipage}{\linewidth}
  \centering
  \includegraphics[width=0.5\linewidth]{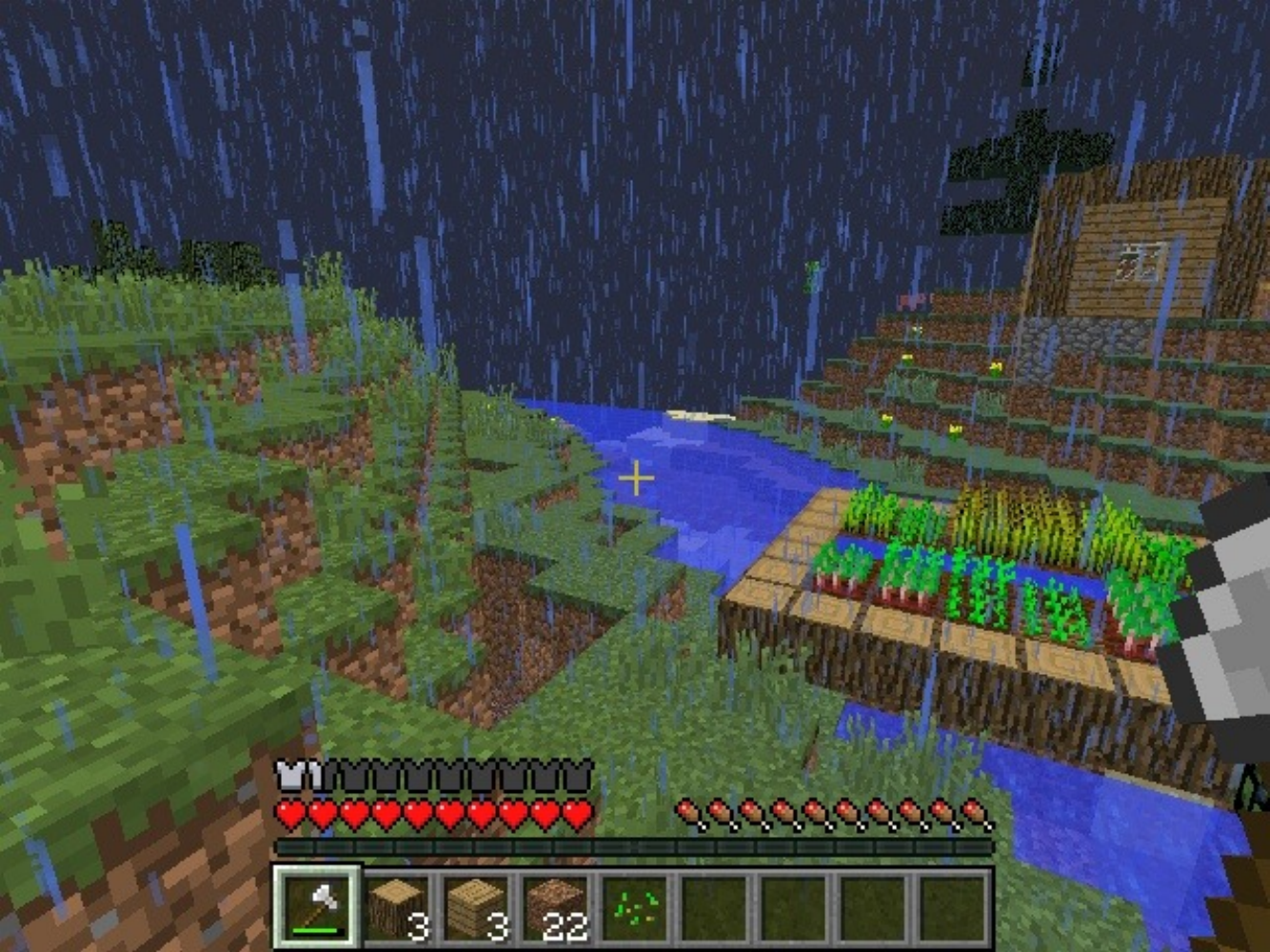}
  \label{fig:app2}
\end{minipage}

\vspace{5pt}

\textbf{Question:} What time of day is it in the game?

\vspace{2pt}

\textbf{GPT-4o:} It is difficult to determine the exact time of day in the game due to the rainy weather, which can create a darkened sky. However, based on the darkness of the surroundings, it appears to be night or close to night. Rain in Minecraft makes it harder to see the exact position of the sun or the moon, but the overall dim lighting suggests it is either late evening or night time.

\vspace{2pt}

\textbf{Qwen2-VL:} It is daytime in the game. The bright light and the visible sky indicate that it is daytime.

\vspace{2pt}

\textbf{\model-Qwen2-VL:} The time of day in the game is night. This can be determined by the dark sky and the presence of rain in the screenshot. In Minecraft, night begins when the sun sets and the moon rises. During nighttime, the sky darkens, and if it's raining, you'll see rain particles falling from the sky. Additionally, night is a more dangerous time in Minecraft as hostile mobs spawn more frequently.

\vspace{5pt}

\end{minipage}
\end{center}
\end{prompt}

\subsection{Spatial Grounding Evaluation}
For Spatial Grounding Evaluation, we used 100 GUI data entries and 236 embodied data entries, as seen in \autoref{table:spatial-grounding evaluation}. We required the model to output the points location of a specified object in the image. If there were no points, the bounding box would be used as a substitute. The output results will be normalized to the range [0, 1000).

\begin{table}[htbp]
\centering
\renewcommand{\arraystretch}{2}
\caption{Summary of spatial grounding evaluation results for visual grounding tasks.}
\label{table:spatial-grounding evaluation}
\begin{tabular}{>{\raggedright\arraybackslash}p{2cm} 
                >{\centering\arraybackslash}p{1cm} 
                >{\centering\arraybackslash}p{2cm} 
                >{\raggedright\arraybackslash}p{4cm} 
                >{\raggedright\arraybackslash}p{4cm}}
\toprule
\textbf{Category} & \textbf{Quantity} & \textbf{Image} & \textbf{Example Question} & \textbf{Example Answer} \\ 
\midrule
GUI & 100 & \multirow{1}{*}{\includegraphics[width=0.1\textwidth]{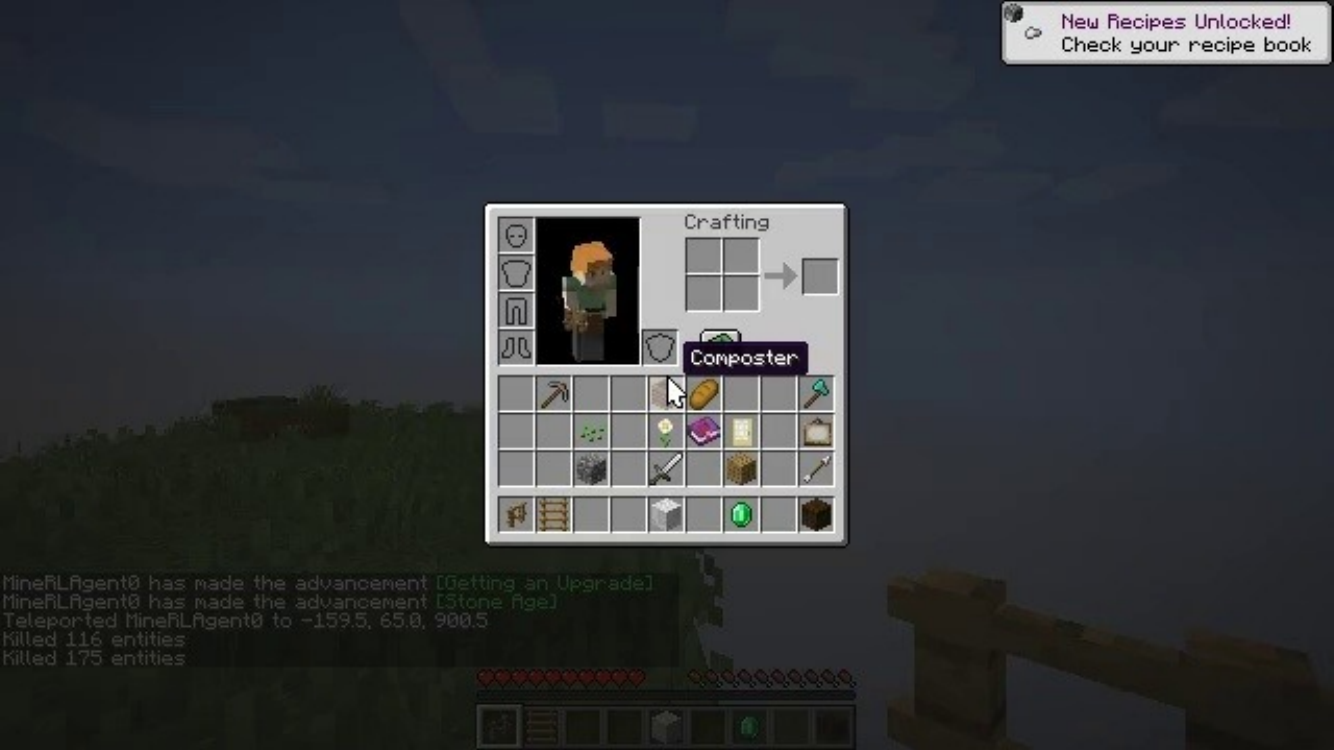}} & Point the wheat\_seeds  & [284,206]  \\ \midrule

Embodied & 236 & \multirow{1}{*}{\includegraphics[width=0.1\textwidth]{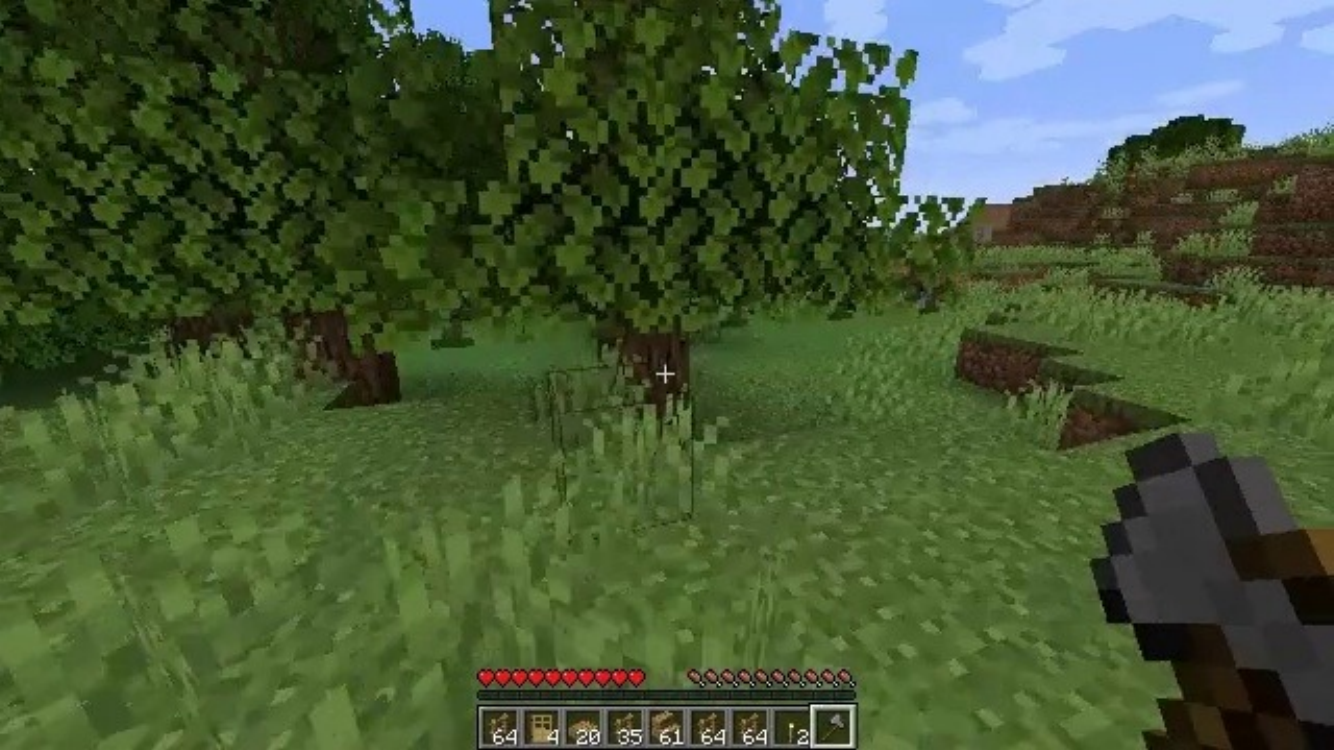}} & Point the oak\_leaves.&  [315,174] \\
\bottomrule
\end{tabular}
\end{table}

\subsection{Evaluation Metric and Result}
    We designed a customized evaluation method to assess the performance of models in answering the questions. For World Knowledge Questions and Visual Understanding Questions, we explore the utilization of LLMs as judges. We selected GPT-4o~\citep{chatgpt}, a state-of-the-art LLM to serve as the judge. The judge model first reviews the responses and compares them to a set of expertly crafted reference answers. Subsequently, the judge assigns a score of correct or incorrect. For visual grounding tasks, we directly score the responses of the evaluated model based on a rule-based approach. Below are the performances of some models we are interested in under our benchmark: \autoref{tab:benchmark}.

\input{tables/benchmark}

\section{Ablation with different Pre-trained VLMs}\label{sec:experiments_vlm_ablation}

In this section, we examine the impact of prior training on a VLMs regarding the robustness of the model's backbone. VLMs vary in their decision-making capabilities due to differences in training data. We highlight this and emphasize the influence of the VLM training architecture on the VLA.

We compare two models, Llava-Next~\citep{llavanext} and Qwen2-VL~\citep{qwen2vl}, which utilize different pretraining datasets and image processing techniques. Their raw VLM performances and post-training results on various auxiliary tasks, along with downstream imitation learning outcomes, are presented in \autoref{tab:model_ablation}.

Both Llava-Next and Qwen2-VL demonstrated more than a 30\% increase in downstream task success rates after undergoing \method post-training. Indicating that improving model performance through visual language post-training is robust across different models.

\input{tables/model_ablation}

%% file: tables/benchmark.tex
\definecolor{lightyellow}{rgb}{1.0, 1.0, 0.6}
\begin{table*}[]
\centering
\resizebox{\textwidth}{!}{
\begin{tabular}{lccccccc}
\toprule
\multirow{2}{*}{Model} & \multirow{2}{*}{Model Size} & \multicolumn{2}{c}{World Knowledge} & \multicolumn{2}{c}{Visual Understanding} & \multicolumn{2}{c}{Visual Grounding} \\ 
\cmidrule(lr){3-4} \cmidrule(lr){5-6} \cmidrule(lr){7-8} 
 & & Acc & Rank & Acc & Rank & Acc & Rank \\ 
\midrule
GPT-4o~\citep{gpt-4} & - & 96.6 & 1 & 76.7 & 1 & - & - \\
GPT-4o-mini~\citep{gpt-4} & - & 75.9 & 2 & 62.8 & 4 & - & - \\
\midrule
Llava-Next~\citep{llavanext} & 8B & 19.0 & 8 & 41.9 & 10 & - & - \\
Molmo-d-0924~\citep{molmo} & 7B & 12.1 & 10 & 58.1 & 5 & 24.8 & 3 \\
Llama-3.2~\citep{llama3.2} & 11B & 20.7 & 7 & 44.2 & 9 & - & - \\
Qwen2-VL~\citep{qwen2vl} & 7B & 17.3 & 9 & 46.5 & 7 & 16.6 & 5 \\
\midrule
\textbf{Qwen2-VL~(Knowledge)} & 7B & 65.5 & 5 & 46.5 & 7 & 16.6 & 5 \\
\textbf{Qwen2-VL~(Vision)} & 7B & 62.1 & 6 & 65.1 & 3 & 19.8 & 4 \\
\textbf{Qwen2-VL~(Grounding)} & 7B & 67.2 & 4 & 51.2 & 6 & 63.6 & 2 \\
\textbf{\model-Qwen2-VL} & 7B & 70.7 & 3 & 76.7 & 1 & 88.0 & 1 \\
\bottomrule
\end{tabular}
}
\caption{We compared the performance of various VLMs using our benchmark, including commercial large models (GPT-4 and GPT-4-mini~\citep{chatgpt}), open-source models (Llava-Next~\citep{llavanext}, Molmo-d-0924~\citep{molmo}, Llama-3.2~\citep{llama3.2}, and Qwen2-VL~\citep{qwen2vl}), as well as \model. The results demonstrate that our method significantly enhances the core capabilities of these models, although there remains a gap when compared to state-of-the-art models.}
\label{tab:benchmark}
\end{table*}

%% file: tables/model_ablation.tex
\begin{table*}[htbp]

\centering

\label{tab:model_ablation}

\begin{tabular}{l cccc}

\toprule

Model & Diamond Sword & Ladder & Cooked Beef & Iron Ingot \\

\midrule

Qwen2-VL(raw) & 0.53 & 0.40 & 0.03 & 0.10 \\

Qwen2-VL-7B (one-stage) & 0.10 & 0.40 & 0.07 & 0.13 \\

\method-Qwen2-VL & 0.83 & 0.63 & 0.77 & 0.70 \\

\bottomrule

\end{tabular}
\caption{Success rates comparing different training paradigms on selected Minecraft tasks. 'raw' refers to the Qwen2-VL baseline fine-tuned directly. 'RT-2 (1-stage co-finetuning)' corresponds to the Qwen2-VL-7B (one-stage) baseline. 'Act-VLP (3-stage post-training)' represents our \method-Qwen2-VL approach.}

\end{table*}